\pdfoutput=1
\documentclass[10pt,twocolumn,letterpaper]{article}

\usepackage{iccv}
\usepackage{times}
\usepackage{epsfig}
\usepackage{graphicx}
\usepackage{amsmath}
\usepackage{amssymb}

\usepackage[accsupp]{axessibility}  

\usepackage{paralist}
\usepackage{pifont}
\usepackage{multirow}
\usepackage{booktabs}
\usepackage{caption}
\usepackage[font=normal]{subfig}
\usepackage{tabularx}
\usepackage{makecell}

\usepackage{xcolor}
\usepackage{color, colortbl}
\usepackage{amssymb}

\usepackage[symbol]{footmisc}

\newcommand{\figref}[1]{Figure~\ref{#1}}
\newcommand{\tabref}[1]{Table~\ref{#1}}

\newcommand{\tabspace}{\vspace{-4mm}}
\newcommand{\tabcspace}{\vspace{-2mm}}
\newcommand{\figcspace}{\vspace{-2mm}}
\newcommand{\figspace}{\vspace{-4mm}}

\newcommand{\Paragraph}[1]{\noindent\textbf{#1}}
\newcommand{\paren}[1]{\left( #1 \right)}

\usepackage[pagebackref=true,breaklinks=true,colorlinks,bookmarks=false,citecolor=blue,linkcolor=blue]{hyperref}

\iccvfinalcopy 

\def\iccvPaperID{6678} 

\ificcvfinal\fi

\begin{document}

\title{C2N: Practical Generative Noise Modeling for Real-World Denoising}

\author{
Geonwoon Jang\thanks{\textit{Authors contributed equally.}}
~~~~~~~~
Wooseok Lee\footnote[1] ~~~~~~~~
Sanghyun Son ~~~~~~~~
Kyoung Mu Lee\\
ASRI, Department of ECE, Seoul National University\\
{\tt\small \{onwoono, adntjr4\}@gmail.com, \{thstkdgus35, kyoungmu\}@snu.ac.kr}
}

\maketitle
\ificcvfinal\thispagestyle{empty}\fi

\begin{abstract}
    Learning-based image denoising methods have been bounded to situations where well-aligned noisy and clean images are given, or samples are synthesized from predetermined noise models, e.g., Gaussian.
    While recent generative noise modeling methods aim to simulate the unknown distribution of real-world noise, several limitations still exist.
    In a practical scenario, a noise generator should learn to simulate the general and complex noise distribution without using paired noisy and clean images.
    However, since existing methods are constructed on the unrealistic assumption of real-world noise, they tend to generate implausible patterns and cannot express complicated noise maps.
    Therefore, we introduce a Clean-to-Noisy image generation framework, namely C2N, to imitate complex real-world noise without using any paired examples.
    We construct the noise generator in C2N accordingly with each component of real-world noise characteristics to express a wide range of noise accurately.
    Combined with our C2N, conventional denoising CNNs can be trained to outperform existing unsupervised methods on challenging real-world benchmarks by a large margin.
\end{abstract}

\section{Introduction}
Image denoising aims to remove unintended signals from a given noisy observation.
The task has been considered as one of the fundamental vision problems and handled by numerous studies~\cite{NLmeans, BM3D, WNNM}.
While recent deep convolutional neural networks~(CNNs) have achieved promising performance~\cite{DnCNN, FFDNet, CBDNet, DDFN, GradNet_DN}, several challenges prevent them from being used for practical applications.
A primary limitation of the learning-based approaches is that they are usually data-driven, where training on a specific dataset does not guarantee generalization to a real-world scenarios~\cite{Gau_meets_Real, CBDNet}.

\begin{figure}[t]
    \renewcommand{\wp}{0.245 \linewidth}
    \centering
    \subfloat[Clean]{\includegraphics[width=\wp]{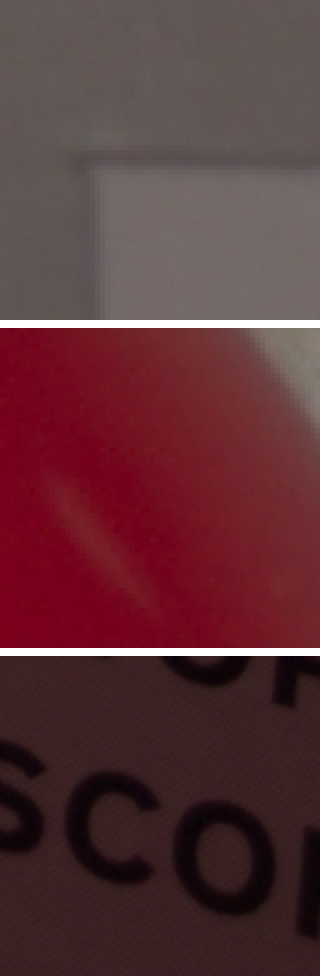}}
    \hfill
    \subfloat[GT]{\includegraphics[width=\wp]{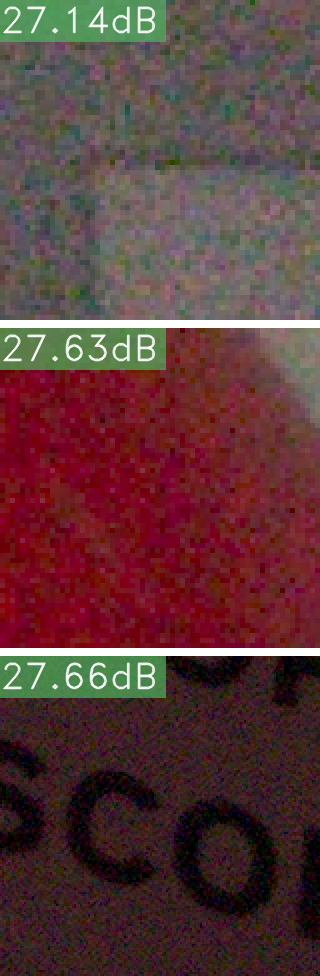}}
    \hfill
    \subfloat[\textbf{C2N}]{\includegraphics[width=\wp]{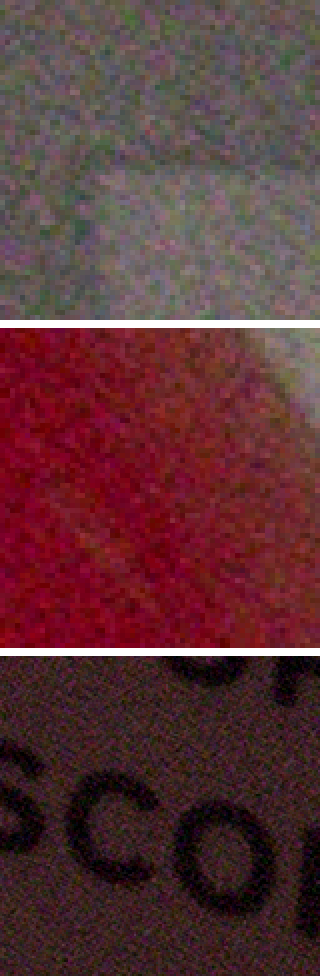}}
    \hfill
    \subfloat[\textbf{DIDN}~\cite{DIDN}]{\includegraphics[width=\wp]{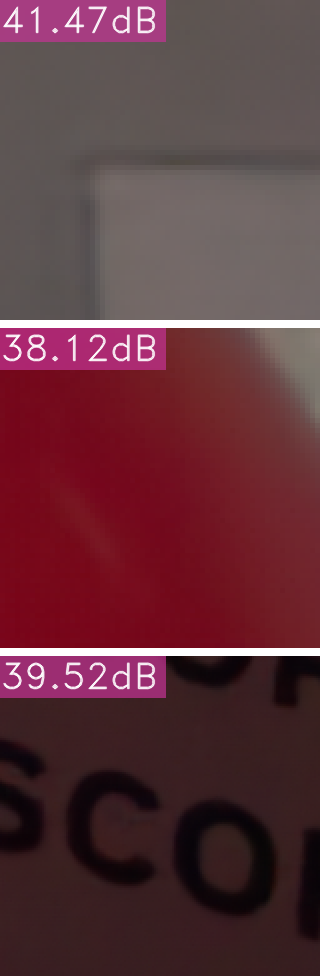}}
    \figcspace
    \caption{
    \textbf{Examples of generated and denoised image from our proposed method.}
    (a) Clean image, (b) Ground truth noisy image,
    (c) Generated noisy image from the proposed C2N,
    (d) Denoising results of DIDN~\protect\cite{DIDN} trained on the images generated by our C2N.
    Our C2N can accurately imitate the real noise without using paired examples.
    }
    \label{fig:results-intro}
    \figspace
\end{figure}

The noise from a typical camera pipeline is different to the conventional assumption for ideal noise in several aspects.
For instance, a widely-used Additive White Gaussian Noise (AWGN) formulation assumes that the term is signal-independent~\cite{GAN2GAN, N2V}, while real-world noises are not.
Therefore, it is difficult to generalize a denoising algorithm toward real-world images when the model is trained on synthetic examples.
As an alternative, few studies have collected well-aligned noisy and clean image pairs in the wild~\cite{SIDD, DND} so that the following denoising methods can be trained in a supervised manner.
While such an approach is an effective way to deal with real-world noise, it remains challenging to acquire large-scale pairs due to several practical issues.
Recent self-supervised approaches~\cite{N2V, N2S} deal with the limitations by using noisy samples as the target output, whose supervision would lead a model to estimate the actual value on average.
However, they usually leverage some statistical properties of the noise, which are insufficient to represent the real-world cases~\cite{SIDD, DND}.

On the other hand, generation-based approaches~\cite{GCBD, UIDNet} have been proposed to deal with challenging real-world situations.
These approaches first train a noise generator~\cite{GAN} on target noisy images to produce pseudo-noisy images paired to other clean images, which are used later to train a denoising model.
Following the successful results of the earlier methods~\cite{GCBD, GAN2GAN} for synthetic noise reduction, recent attempts~\cite{NoiseFlow, UIDNet} to apply this method to real-world noise have been introduced.
Still, in a situation where no paired clean images are given to the target noisy images, a generation-based method that successfully imitates real-world noise has not been suggested.

In this paper, we introduce C2N(Clean to Noisy), a novel generative noise modeling framework trained without any paired data.
The C2N can learn a variety of complex noise distributions successfully and generates accurate noisy images from arbitrary clean images.
We achieve state-of-the-art performance by training existing denoising models with the generated pairs from our C2N framework.
Our contributions is summarized as follows:
\begin{compactitem}[$\bullet$]
    \item We propose a novel noise generator with explicit modules to express noise terms of according characteristics, enabling the generator to imitate more accurate real-world noise.
    \item Our C2N framework successfully simulates target noise distribution without any handcrafted formulations or unrealistic assumptions.
    \item With the data pairs generated by C2N, we train denoising models that outperform state-of-the-art unsupervised methods for denoising real photographs.
    The test code and pretrained models are publicly available at: \url{https://github.com/onwn/C2N}.
\end{compactitem}

\section{Related Works}
\Paragraph{Deep Image Denoising.}
After the DnCNN~\cite{DnCNN} model has achieved a significant performance gain over traditional methods \cite{NLmeans, BM3D}, CNN-based methods have become mainstream in the image denoising area.
The FFDNet~\cite{FFDNet} model uses given noise level maps to remove AWGN with spatially varying noise levels effectively.
Using large models and complex architectures, the performance of denoising models can be improved by extracting rich features from the input noisy image \cite{DDFN, FOCNet}.
While such methods successfully erase AWGN with state-of-the-art performances, they still require numerous training pairs that contain the exact target noise distribution.
Therefore, the real-world denoising task has remained challenging since the desired noise model is unknown without appropriate training examples.

\Paragraph{Deep Denoising of Real-World Noise.}
If enough amount of training samples are given, it is straightforward to train the methods mentioned to function properly.
For such purpose, Anaya and Barbu~\cite{Renoir} acquire the Renoir dataset where clean samples are synthesized from a sequence of low-ISO images.
However, spatial misalignment and remaining noise in training pairs make it challenging to use the dataset for practical purpose.
The DND~\cite{DND} dataset post-processes low-ISO images to align their spatial contents and illuminations with high-ISO counterparts.
The SIDD~\cite{SIDD} dataset captures noisy images under various lighting conditions with five different smartphone cameras.
With these real-world noise datasets, denoising models with various attention modules \cite{GRDN, RIDNet}, multi-scale resizing in features \cite{U-net, multi_wavelet, DIDN}, or use of self-similarity in images \cite{2017NLnet, NLRN, DeepProxies_BM3D} manage to remove complex real noises.

Nevertheless, it is difficult to collect large-scale real-world dataset for our specific purpose.
To remove noise even when the accurate distribution of it is not given(\eg bling denoising), the CBDNet~\cite{CBDNet} includes a part that performs noise level estimation based on heteroscedastic Gaussian noise model~\cite{heteroGau}.
It is intended to also operate on real noisy images, receiving indirect supervision from training the denoiser.
The Path-Restore~\cite{Path-Restore} dynamically selects an appropriate restoration path for each region of an input image.
The self-supervised denoising methods \cite{N2V, N2S} use only individual noisy images for training and estimates a pixel value of its input noisy image itself, where the value of that location is masked-out as 'blind spot'.
Since these methods require the noise to satisfy strong statistical assumptions, \cite{StructN2V, laine2019high} modify these according to the prior knowledge of noisy images.
Recently, the AINDNet~\cite{AINDNet} apply adaptive instance normalizing method to deal with varying noise levels.
And the Noise2Blur~\cite{N2B} performs an additional procedure to preserve image details after training a model with blurred labels.

\Paragraph{Noise Generation-Based Denoising.}
The generation-based methods~\cite{GCBD, GAN2GAN, NoiseFlow, UIDNet} usually adopt a two-stage pipeline for the denoising problem rather than using a single model.
First, a noise generator is learned in an unsupervised fashion~\cite{GAN} to simulate the distribution of given real noisy examples so that any clean images can be mapped to pseudo-noisy data.
The GCBD~\cite{GCBD} is the first generation-based approach for deep blind image denoising.
Later, the GAN2GAN~\cite{GAN2GAN} method leverages better noisy-patch-extraction, generating more realistic noisy samples to train the following denoising model in an N2N~\cite{N2N} manner.
While the methods above are limited to show promising results for signal-independent and synthetic noise only, the Noise Flow~\cite{NoiseFlow} proposes a formulation to imitate challenging real-world noise.
By leveraging the normalizing flow formed of invertible transforms, it can precisely learn the shift of distribution between synthetic and desired noise maps.
Although the model successfully imitates in-camera noise occurring pipeline, true noisy and clean image pairs are required to get the correct noise distribution, which is not practical.
The DANet~\cite{DANet} also learns the mapping of denoiser and generator by comparing them with the true noisy-to-clean joint distribution, but it still needs paired images to learn such mapping. 
Recently, the UIDNet~\cite{UIDNet} model handles the unpaired generative noise modeling with the sharpening technique for better noise separation.
Furthermore, the NTGAN~\cite{NTGAN} method demonstrates that noise maps synthesized with a given camera response function (CRF) can be used for the following denoising network.
Unlike the existing generation-based denoising methods, we introduce a novel noise generator trained without any paired data or heuristic methods.

\section{Method}
\subsection{Complexity of Real-World Noise}
\label{sec:char_RealN}
To generate realistic noise with CNNs, it is necessary to understand the properties and statistical behavior of real-world noise.
Due to several physical limitations, the noise occurs from various sources, including electronic sensors, in-camera amplifiers, photon noise, quantization, and compression artifacts~\cite{1998camera}.
Combining all these factors, the pixel-wise noise term $n$ is mixed with an underlying clean signal $x$, resulting the noisy observation $y$ as follows:
\begin{equation} \label{eq:noisy_and_clean}
    y = x + n.
\end{equation}

In conventional deep denoising methods~\cite{DnCNN, FFDNet}, the noise term $n$ is usually simplified as an ideal Additive White Gaussian Noise (AWGN), i.e., $n \sim \mathcal{N}(0,\sigma^2)$, where 
$\sigma$ denotes the standard deviation.
On the other hand, the photon noise is signal-dependent, where Poisson distribution can be used to simulate the case and be approximated as a Gaussian distribution with signal-dependent variance~\cite{heteroGau}.
The heteroscedastic Gaussian noise is defined as follows:
\begin{equation} \label{eq:heteroGau}
    n \sim \mathcal{N}(0, \sigma_s^2 x + \sigma_c^2),
\end{equation}
where $\sigma_s$ and $\sigma_c$ are hyperparameters for signal-dependent and signal-independent term.
While the noise model in \eqref{eq:heteroGau} can provide a proper approximation of the realistic noise~\cite{CBDNet, NTGAN} to some extent, several studies~\cite{1998camera, ELD} have demonstrated that the real-world cases appear to be much more complicated.
In addition, the physical limitations of in-camera electronic devices and in-camera compression pipeline make the noise term to exhibit random spatial pattern~\cite{2009img_proc}.
Such property leads the noise term $n$ and its local neighbors to be spatially correlated, making the precise modeling more challenging.

To deal with the problem without using paired data, previous approaches~\cite{CBDNet, NTGAN} construct noise maps using a synthetic noise model and known camera response function (CRF).
However, such handcrafted features prevent those methods from being generalized toward realistic configurations.
Therefore, we adopt a learning-based method, namely C2N, to simulate the real-world noise rather than using some handcrafted formulations.
Our framework fully utilizes the advantage of unsupervised learning to simulate the comprehensive real-world noise, with novel design components and objective terms.

\subsection{Learning to Generate Pseudo-Noisy Images} \label{use_train_C2N}
A denoising network $F$ aims to reconstruct the underlying clean signal $x$ from a given noisy observation $y$ in \eqref{eq:noisy_and_clean}.
When enough training pairs are available, the model can be trained in a supervised manner to estimate the clean signal.
However, in real-world scenarios, it is challenging to acquire \emph{ideal} clean images well-aligned to the training noisy images, even with complicated post-processing~\cite{SIDD}.
Hence, our C2N framework first trains a generator to simulate the target noise distribution.
Then, the following denoising model $F$ can be learned on the generated noisy examples.
\begin{figure}[t]
    \renewcommand{\wp}{0.48 \linewidth}
    \centering
    \subfloat[Training generator]{\includegraphics[width=\wp] {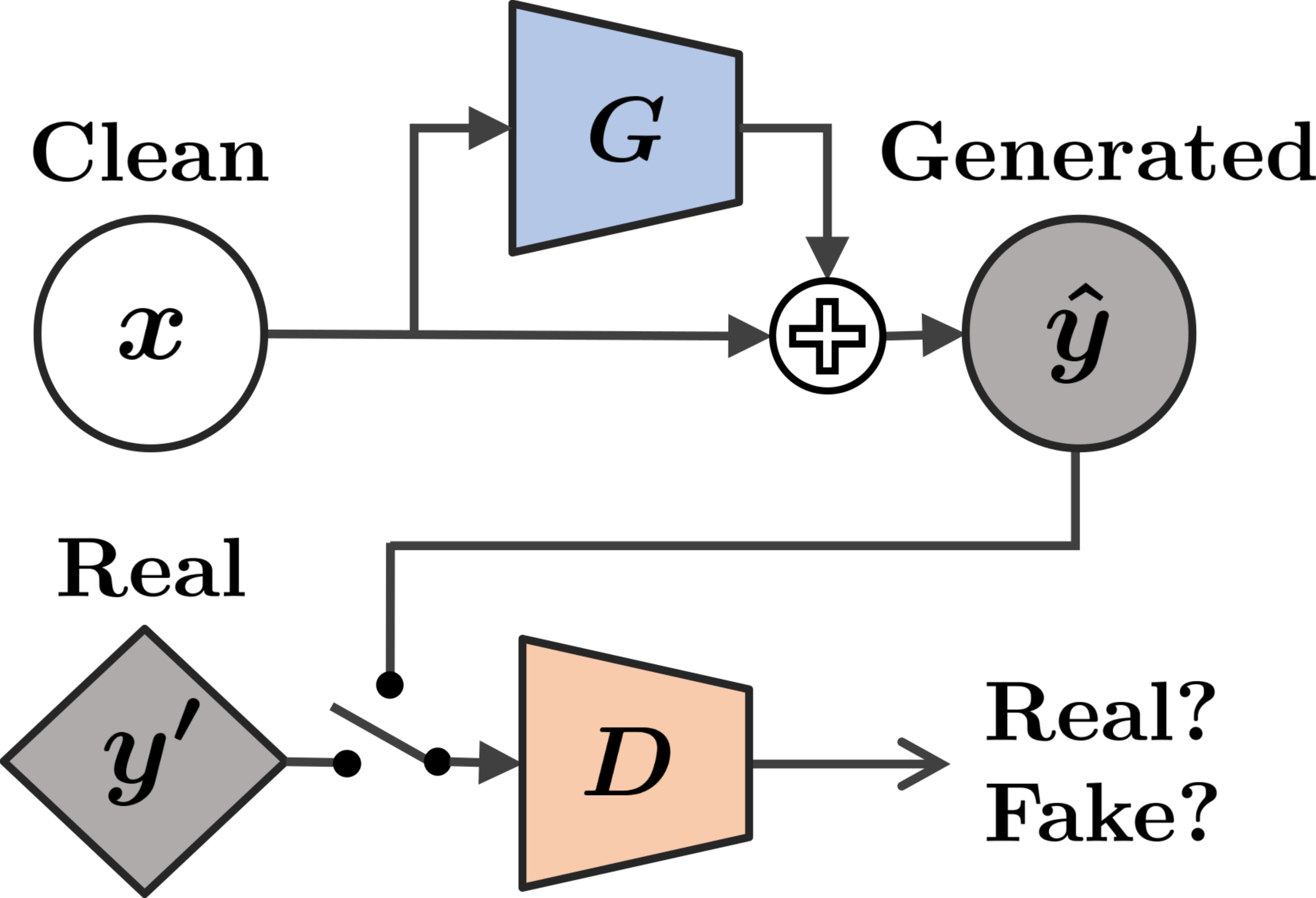}}
    \hfill
    \subfloat[Denoising]{\includegraphics[width=\wp] {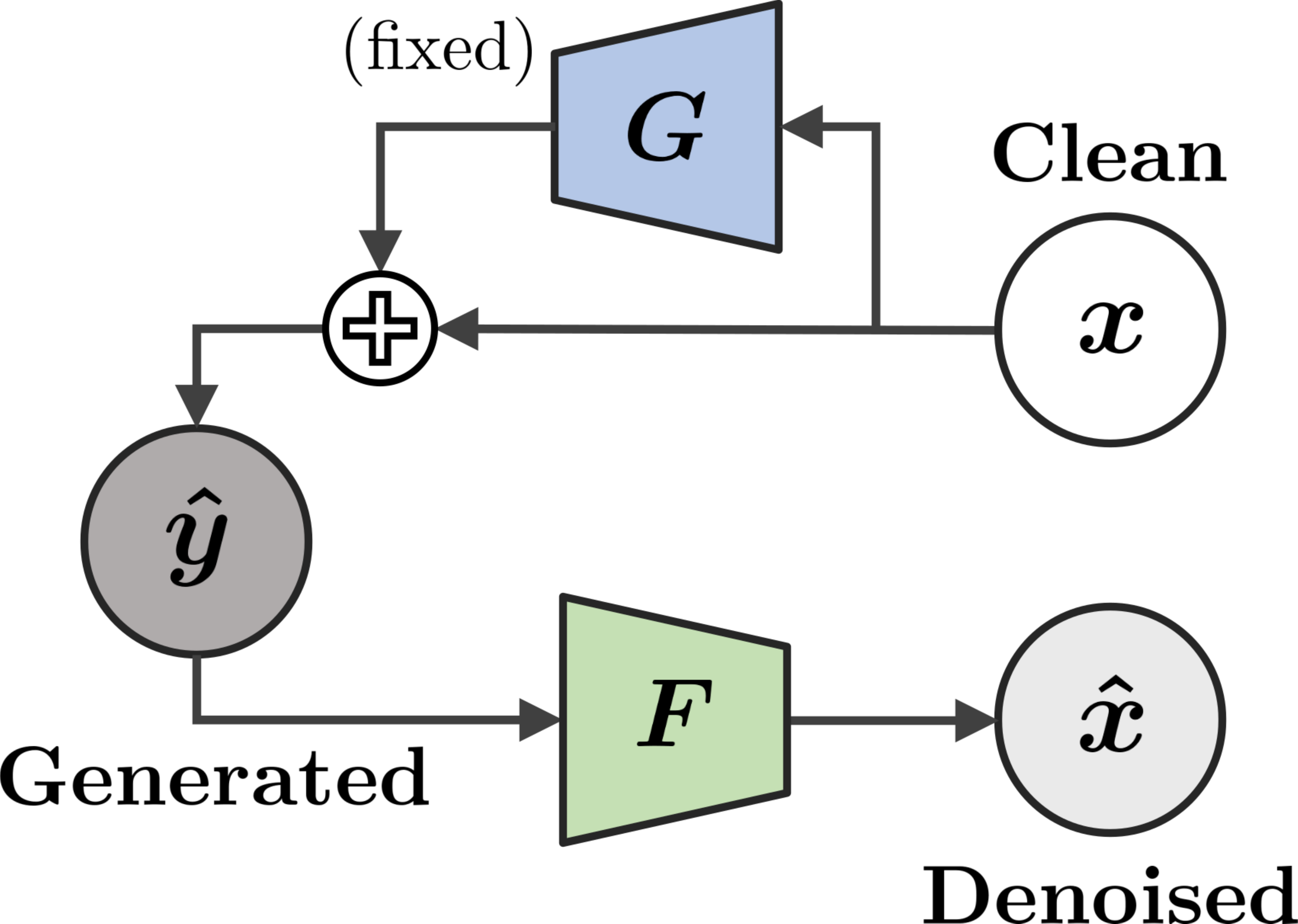}}
    \\
    \figcspace
    \caption{
        \textbf{Our two-step pipeline for real-world denoising.}
        (a) Our method first learns to generate samples from target noise distribution, with a noise generator $G$.
        We used clean image $x$ and noise image $y'$ to be unpaired. 
        (b) Secondly, using the generated pairs, we train a denoising model $F$.
    }
    \label{fig:two-step}
    \figspace
\end{figure}
\figref{fig:two-step} shows the two-step pipeline of our method.

Our noise generator network $G$ is designed to synthesize a realistic noise map $\hat{n}$ for a given clean image $x$ to produce the pseudo-noisy image $\hat{y}$ as follows:
\begin{equation} \label{eq:G}
    \hat{y} = x + \hat{n} = x + G \left( x, r \right),
\end{equation}
where $r$ is a random vector to reflect the stochastic behavior of noise according to the conditions of each scene.
We sample 32-dim random vector from $\mathcal{N}(0,1^2)$ and spatially replicate through all pixel positions of $x$, similar to the GAN applications \cite{cGAN, T2I}.

\begin{figure*}[t]
    \centering
    \includegraphics[width=\linewidth]{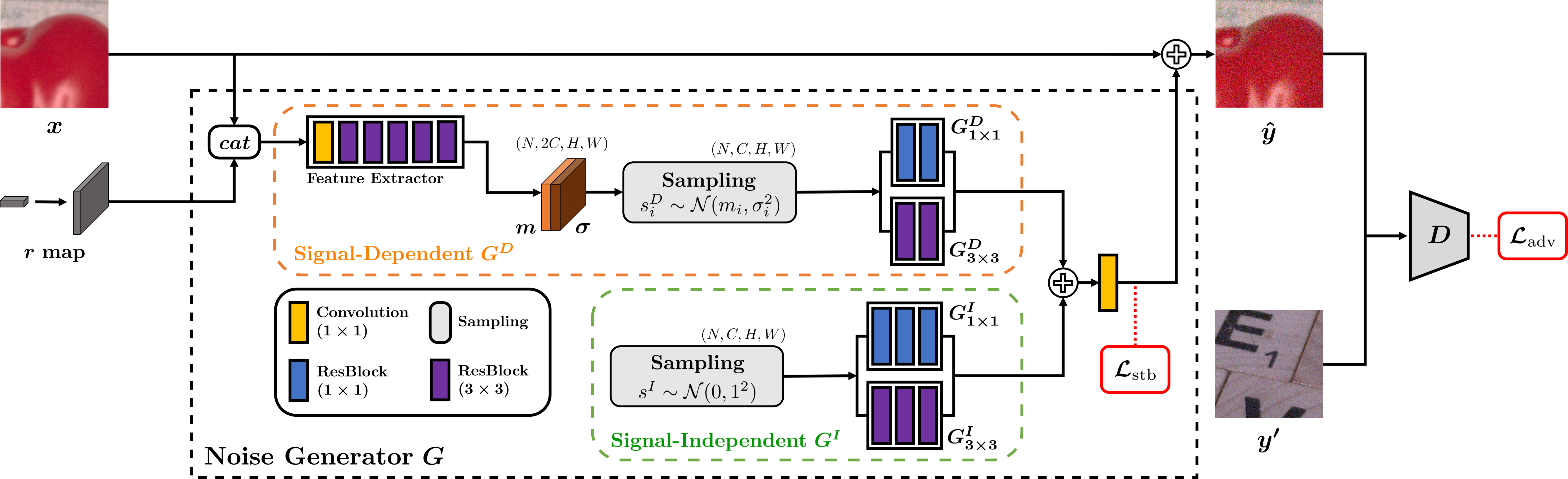}
    \figcspace
    \vspace{-3mm}
    \caption{
    \textbf{Overview of C2N framework.}
    The noise generator $G$ architecture of our C2N and its components are shown.
    $r$ map denotes the spatially replicated input random vector $r$.
    Initial noise maps $s^I$ and $s^D$ are sampled in the signal-independent part $G^I$ and signal-dependent part $G^D$ of the generator, respectively.
    }
    \label{fig:architecture}
    \figspace
\end{figure*}

Simultaneously, we train a discriminator network $D$ to distinguish whether a given noisy image is synthesized from our generator $G$ or sampled from the real-world dataset.
The two networks $G$ and $D$ can be optimized in an adversarial way~\cite{GAN}, using the Wasserstein distance~\cite{WGAN-GP} as follows:
\begin{equation}
\begin{split} \label{eq:L_adv}
    \mathcal{L}_\text{adv}(D,G) &= \mathbb{E}_{y' \sim P_{N}}[D(y')] \\
    &- \mathbb{E}_{x \sim P_{C}, r \sim P_r}[D(x+G(x, r))] \\
    &+ \lambda \mathbb{E}_{x_\delta \sim P_\delta}[( \left\| \nabla_{x_\delta} D(x_\delta) \right\|_2- 1)^2],
\end{split}
\end{equation}
where the term is minimized with respect to $G$ and maximized with respect to $D$.
$P_{N}$ and $P_{C}$ denote the distribution of the real-world noisy and clean images, respectively.
The real noisy image $y'$ is sampled from $P_{N}$. 
Prime notation on $y$ denotes that we use noisy image unpaired to clean image $x$.
$P_r$ is the distribution of random vector $r$.
For stable learning, we adopt the gradient penalty~\cite{WGAN-GP}, which is weighted by a hyperparameter $\lambda=10$.
The term $x_\delta \sim P_\delta$ is one of the internal dividing points between generated and real images.

The significant advantage of the proposed C2N framework is that generator in the C2N can synthesize realistic noise without adopting handcrafted features.
However, the generated noise $\hat{n}$ may bias the image color and negatively affect the overall framework if our C2N is trained without any constraints.
To deal with the case, we additionally define a stabilizing loss term $\mathcal{L}_\text{stb}$, which is defined as follows:
\begin{equation} \label{eq:L_stb}
    \mathcal{L}_\text{stb} = {\frac{1}{N}} \sum_{c} \left\| \sum_{i \in B} \hat{n}_{i,c} \right\|_1,
\end{equation}
where $N$ denotes the number of pixel $i$ in mini-batch $B$ and $c$ is index of each color channels.
By minimizing the stabilizing loss, the channel-wise average of the generated noise approaches zero, which prevents the color-shifting problem.
Since we do not take a mean over a single sample or local area, the generated noise may have varying nonzero local means depending on the underlying signal.
Combining our two loss terms \eqref{eq:L_adv} and \eqref{eq:L_stb}, we optimize the total loss $\mathcal{L}_{G} = \mathcal{L}_\text{adv} + w_\text{stb} \mathcal{L}_\text{stb}$, where $w_\text{stb}=0.01$.

\subsection{C2N Architecture} \label{generator}
The previous generation-based approaches have limited their scopes to signal-independent~\cite{GCBD} or spatially uncorrelated~\cite{NoiseFlow} noise terms.
On the other hand, we construct a new generator architecture to represent diverse and complex noise distributions discussed in Section~\ref{sec:char_RealN}.
We gradually implement several design components to express more general properties of real-world noise.
\figref{fig:architecture} shows the overall architecture of our C2N framework.

\paragraph{Signal-Independent Pixel-Wise Transform.}
The noise generation process can be formulated as a nonlinear mapping from an initial random noise map to the desired noise map.
To generate signal-independent noise, we sample the initial random noise map $s^I$ from the standard normal distribution $\mathcal{N} \paren{ 0, 1^2 }$.
We construct a signal-independent pixel-wise noise transformation module $G^I_{1 \times 1}$ to simulate spatially i.i.d. noise.
$G^I_{1 \times 1}$ module consists of 3 modified residual blocks~\cite{ResNet} for low-level vision problems~\cite{EDSR} with $1 \times 1$ convolutional layers without batch-normalization.

\paragraph{Signal-Dependent Pixel-Wise Transform.}
To express signal-dependent noise, our noise generator should extract useful features from the clean input image.
However, it is not desirable that the generated noise map $\hat{n}$ is deterministically generated from the given clean image.
To effectively represent the signal-dependent noise term, we sample initial noise map $s^D$ from position-wise normal distribution which has mean $m_i$ and standard deviation $\sigma_i$ at each position $i$ from convolutional features.
Input clean image and random vector $r$ are transformed to convolutional features through the feature extractor.
Feature extractor consists of 5 residual blocks after $1 \times 1$ convolutional layer without using any pre-trained model.
We then sample the initial noise map $s^D_i$ at each position $i$, and apply another pixel-wise noise transforms $G^D_{1 \times 1}$ which consists of 2 residual blocks with $1 \times 1$ convolutional layer.
For the sampling of $s^D_i$, we used reparameterization trick~\cite{stochGrad} to preserve gradients of the parameters, so that the feature extractor can be jointly trained with the entire C2N framework.

\begin{figure}[t]
    \renewcommand{\wp}{0.495\linewidth}
    \renewcommand{\vs}{-3mm}
    \centering
    \subfloat{\includegraphics[width=\wp]{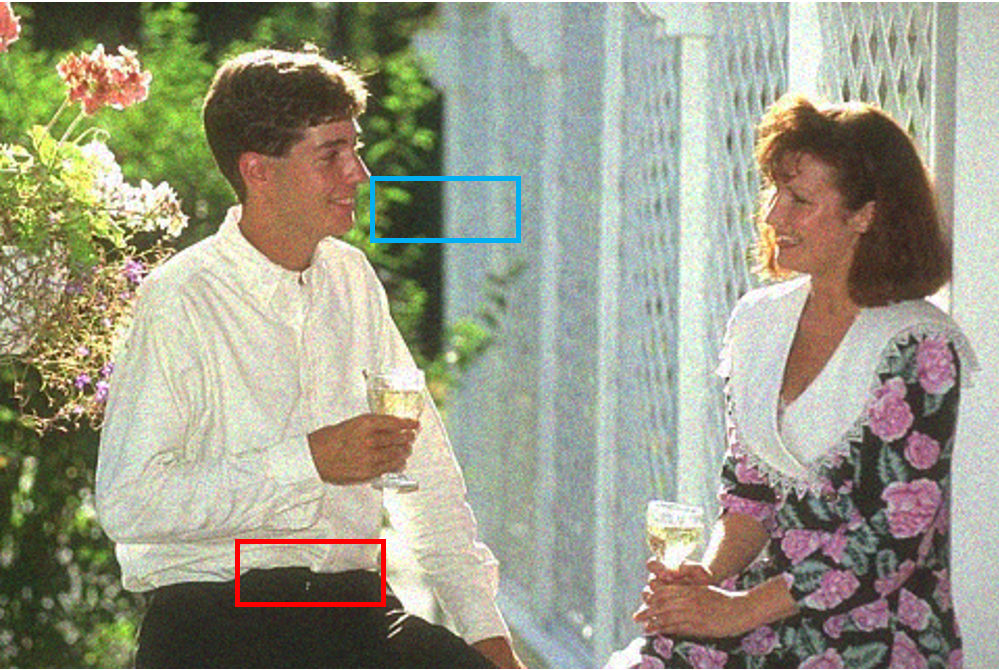}}
    \hfill
    \subfloat{\includegraphics[width=\wp]{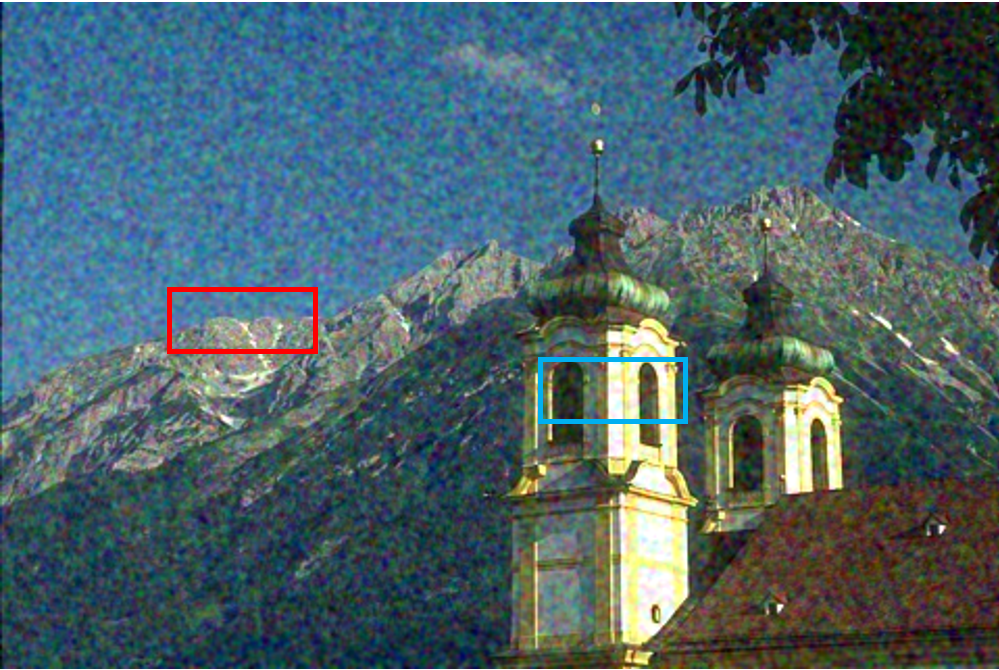}}
    \\
    \vspace{\vs}
    \addtocounter{subfigure}{-2}
    \subfloat[$\mathcal{P}$ (Ground Truth) \label{gt_p}]{\includegraphics[width=\wp]{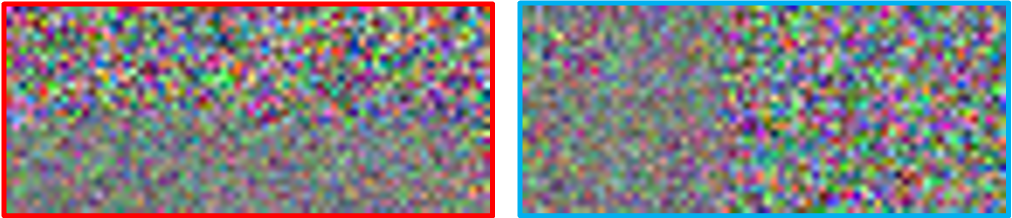}}
    \hfill
    \addtocounter{subfigure}{2}
    \subfloat[$\mathcal{S}$ (Ground Truth) \label{gt_s}]{\includegraphics[width=\wp]{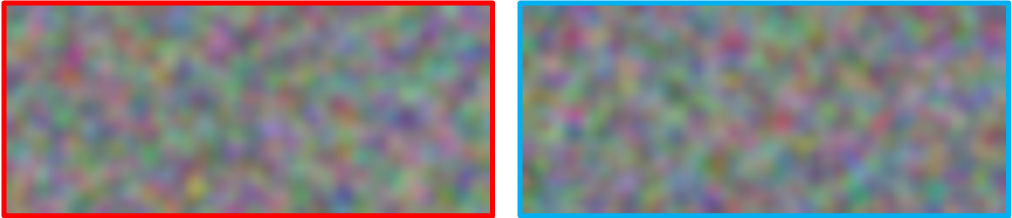}}
    \\
    \vspace{\vs}
    \addtocounter{subfigure}{-3}
    \subfloat[$G^I_{1 \times 1}$ on $\mathcal{P}$ \label{g_i_1x1}]{\includegraphics[width=\wp]{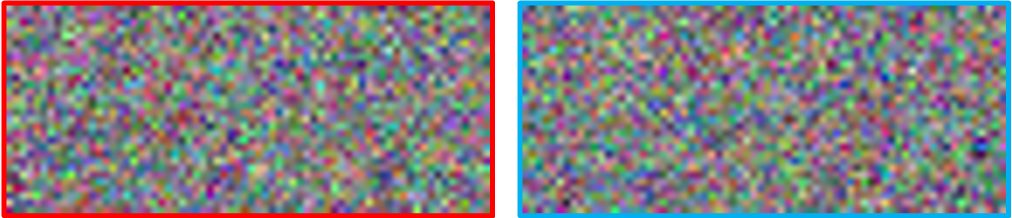}}
    \hfill
    \addtocounter{subfigure}{2}
    \subfloat[$G^I_{1 \times 1} + G^D_{1 \times 1}$ on $\mathcal{S}$ \label{g_i_1x1_g_d_1x1_on_s}]{\includegraphics[width=\wp]{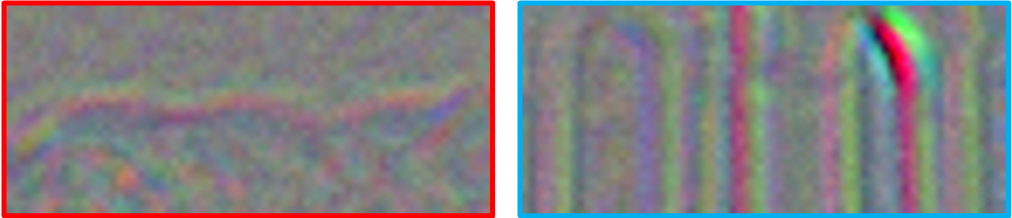}}
    \\
    \vspace{\vs}
    \addtocounter{subfigure}{-3}
    \subfloat[$G^I_{1 \times 1} + G^D_{1 \times 1}$ on $\mathcal{P}$ \label{g_i_1x1_g_d_1x1_on_p}]{\includegraphics[width=\wp]{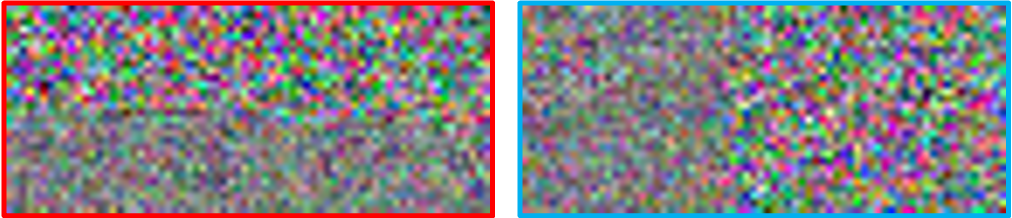}}
    \hfill
    \addtocounter{subfigure}{2}
    \subfloat[$G^I + G^D$ on $\mathcal{S}$ \label{g_i_g_d}]{\includegraphics[width=\wp]{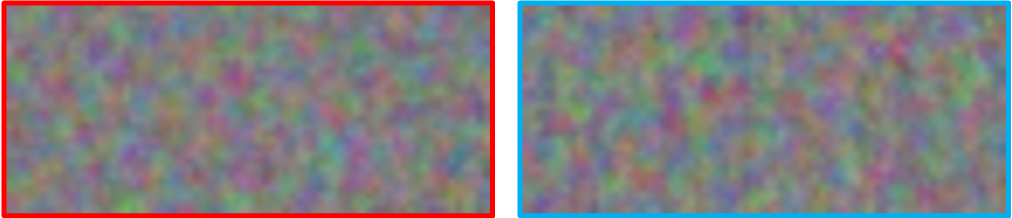}}
    \\
    \figcspace
    \caption{
    \textbf{Generated samples from model ablation study on synthetic noise.}
    (a, d) Synthetic ground-truth noise map of Poisson $\mathcal{P}$ and spatially correlated Gaussian noise $\mathcal{S}$.
    (b, c) Generated residual noise map of the C2N variant trained on the Poisson noise.
    (e, f) Synthesized residual noise on the spatially correlated Gaussian noise.
    We amplify the noise maps in (a-c) by 4 times for clearer comparison.
    }
    \label{fig:ablation-synth}
    \figspace
\end{figure}

\paragraph{Spatially Correlated Transforms.}
A number of existing manual noise models~\cite{heteroGau} and denoising methods assume spatially uncorrelated noise.
On the other hand, C2N handles color sRGB image containing various conversion and compression degradations from in-camera post-processing~\cite{2009img_proc} in end-to-end manner.
To achieve this, We add transforms of $3\times3$ convolution like $G^I_{1 \times 1}$ and $G^D_{1 \times 1}$, which are denoted as $G^I_{3 \times 3}$ and $G^D_{3 \times 3}$.
These $3\times3$ convolutions involve capability for express spatially correlated noise term.
$G^I_{3 \times 3}$ and $G^D_{3 \times 3}$ produce features for the {\em spatially correlated signal-independent noise} and  the {\em spatially correlated signal-dependent noise}, respectively.

Lastly, we add all the noise features transformed through the $G^I_{1 \times 1}$, $G^I_{3 \times 3}$, $G^D_{1 \times 1}$, $G^D_{3 \times 3}$ into one, resulting in a noise feature that integrates the characteristics of noise expressed by each module.
Then we take $1 \times 1$ convolution to the merged feature map to reduce the dimension to the color space, and also to perform a non-linear mapping from the integrated feature to the final noise map.
We set the number of layer channels $C$ as 64, and all the intermediate features from $s^I$ and $s^D$ have the same number of channels.

\subsection{Learning to Denoise with the Generated Pairs}
With the C2N, it is straightforward to optimize the following denoising network $F$.
We first generate pseudo-noisy images $\hat{y}$ from the clean examples $x$ and use the pairs to train a denoising model in a supervised manner~\cite{DnCNN, FFDNet}.
Similar to the previous deep denoising methods \cite{DIDN, GRDN}, we minimize the $L_1$ reconstruction loss which is defined as follows:
\begin{equation} \label{eq:L_rec}
    \mathcal{L}_\text{rec} = \frac{1}{m} \sum_{k=1}^{m} \left\| F(\hat{y}) - x \right\|_1,
\end{equation}
where $\hat{y}$ is pseudo noisy image generated by $\hat{y}=x+G(x,r)$, on $x$ sampled from clean images, and $k$ is index of each images in a mini-batch $B$ of size $m$.

The major advantages of our approach is that our framework is independent to the selection of following denoising architecture.
Previous attempts like \cite{GAN2GAN, UIDNet} train their noise generator and denoising model jointly.
Since the C2N model doesn't get any supervision from the reconstruction loss $\mathcal{L}_\text{rec}$, the generated images are not specialized for certain denoising model.

\section{Experiment}
\subsection{Experimental Setup} \label{setup}
\begin{figure}[t]
    \centering
    \includegraphics[width=\linewidth]{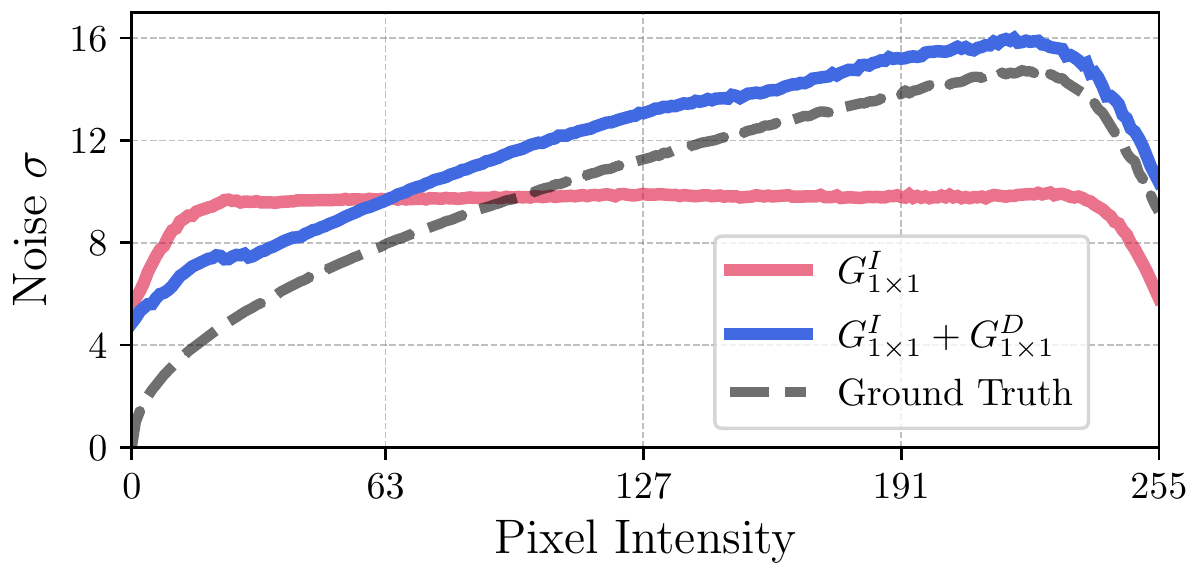}
    \\
    \figcspace
    \caption{
    \textbf{Comparison between generated noise and ground truth Poisson noise.}
    We illustrate pixel noise level with respect to its underlying ground truth signal intensity.
    Noise level is measured on the CBSD68 dataset.
    Distortions around boundary values are due to the clipping of the over-exposed and under-illuminated pixels~\protect\cite{SIDD}.
    }
    \label{fig:plot-ablation-Poi}
    \figspace
\end{figure}

\paragraph{Dataset.}
To train and evaluate our method on synthetic noise, we use clean color images from the BSD500~\cite{BSD} dataset, which consists of 432 training images and 68 test samples.
To train our C2N and denoising model on challenging real-world noise, we leverage the DND~\cite{DND} and the SIDD~\cite{SIDD} dataset captured under realistic environments.
The training split of the SIDD dataset, i.e., SIDD Medium, contains 320 noisy and clean image pairs and the DND dataset contains 50 noisy images.
Since it does not provide any ground truth image for learning on DND, we combine the clean images of the SIDD and noisy images in the DND.
To train our C2N framework, we crop 36,000 patches of size $96 \times 96$ from the noisy and clean image sets and randomly sample them to gather training mini-batches.
For training denoising network, we use only clean patches of the same size and number.
We evaluate the denoising models followed by the C2N with the DND and the SIDD benchmarks.

\paragraph{Implementation Details and Optimization.}
To optimize our C2N framework, we augment all training samples by randomly flipping and rotating them by 90$^\circ$ to construct a mini-batch of size 36.
The Adam~\cite{Adam} optimizer is used with an initial learning rate of $10^{-4}$.
The learning rate is multiplied by a factor of $0.8$ for every $3$ epochs, where a single C2N model is trained over 36 epochs.
Denoising models are optimized using only the generated training pairs from trained C2N generator with mini-batch of size 16.
Same with C2N, the Adam~\cite{Adam} optimizer is used with an intial learning rate of $10^{-4}$ and learning rate is halved for every 4 epochs. 
The training runs for total 16 epochs.
We select the CDnCNN-B~\cite{DnCNN} architecture for denoising as our baseline unless otherwise stated.

\subsection{Model Analysis of C2N on Synthetic Noise} \label{model_analysis-synth}

\begin{table}[t]
    \centering
    \renewcommand{\arraystretch}{0.9}
    \setlength{\belowcaptionskip}{4pt}
    \begin{tabularx}{\linewidth}{p{1.2cm} c >{\centering\arraybackslash}X>{\centering\arraybackslash}X>{\centering\arraybackslash}X}
        \toprule
        \multicolumn{2}{c}{\makecell{Training}} & \multicolumn{3}{c}{Test Noise Level} \\
        \multicolumn{2}{c}{\makecell{Data}} & $\sigma=15$ & $\sigma=25$ & $\sigma=50$ \\
        \midrule
        \multirow{3}{*}{\makecell{Synthetic}} & $\sigma=15$ & \textbf{33.48} & 27.24 & 18.30 \\
        & $\sigma=25$ & 31.39 & \textbf{30.68} & 20.88 \\
        & $\sigma=50$ & 27.82 & 27.81 & \textbf{27.14} \\
        \midrule
        \multicolumn{2}{c}{C2N} & \textbf{32.96} & \textbf{30.51} & \textbf{27.09} \\
        \bottomrule
    \end{tabularx}
    \tabcspace
    \caption{
        \textbf{Denoising performance on synthetic AWGN.}
        PSNR(dB) is calculated on the CBSD68 dataset.
        We note that the C2N models are trained for each noise level independently.
    }
    \label{tab:result-Gau}
    \tabspace
\end{table}
\begin{figure}[t]
    \setlength{\belowcaptionskip}{4pt}
    \renewcommand{\vs}{-3mm}
    \renewcommand{\wp}{0.24 \linewidth}
    \centering
    \subfloat[GT $x$ \\ ]{\includegraphics[width=\wp]{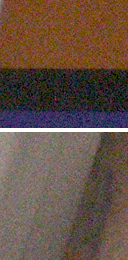}}
    \hfill
    \subfloat[AWGN $\hat{x}$]{\includegraphics[width=\wp]{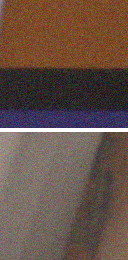}}
    \hfill
    \subfloat[UIDNet $\hat{x}$]{\includegraphics[width=\wp]{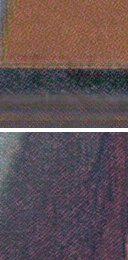}}
    \hfill
    \subfloat[Ours $\hat{x}$]{\includegraphics[width=\wp]{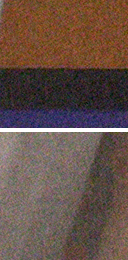}}
    \vspace{\vs}
    \subfloat[GT $n$]{\includegraphics[width=\wp]{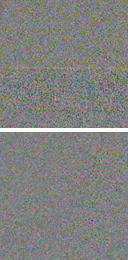}}
    \hfill
    \subfloat[AWGN $\hat{n}$]{\includegraphics[width=\wp]{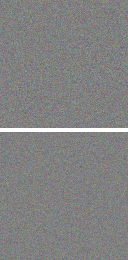}}
    \hfill
    \subfloat[UIDNet $\hat{n}$]{\includegraphics[width=\wp]{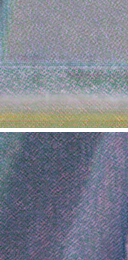}}
    \hfill
    \subfloat[Ours $\hat{n}$]{\includegraphics[width=\wp]{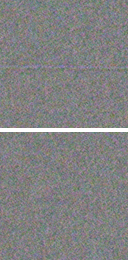}}
    \figcspace
    \caption{
    \textbf{Visual comparison of the generated noisy samples on SIDD dataset.}
    (a) A ground truth noisy image
    (b-d) Generated noisy image of AWGN, UIDNet~\protect\cite{UIDNet} and Ours respectively.
    (e-h) are the residual noise maps of each (a-d).
    }
    \label{fig:comparison-gen}
    \vspace{-6mm}
\end{figure}
\begin{table}[t]
    \centering
    \begin{tabularx}{\linewidth}{>{\centering\arraybackslash}X c c c}
        \toprule
        & AWGN & UIDNet & \textbf{C2N~(Ours)} \\
        \midrule
        KL-divergence & 0.1746 & 0.4417 & \textbf{0.1638} \\
        \bottomrule
    \end{tabularx}
    \tabcspace
    \caption{
        \textbf{KL-divergence between generated noise map and ground truth noise map.}
        The values are calculated using the SIDD validation set.
        We refer the readers to the supplementary materials for more details on the metric.
    }
    \label{tab:results-KL}
    \tabspace
\end{table}
\begin{table}[t]
    \small
    \centering
    \renewcommand{\arraystretch}{0.9}
    \definecolor{Gray}{gray}{0.95}
    \begin{tabularx}{\columnwidth}{>{\centering\arraybackslash}X c >{\centering\arraybackslash}X c >{\centering\arraybackslash}X c}
        \toprule
        $G^I_{1 \times 1}$ & $G^D_{1 \times 1}$ & $G^I_{3 \times 3}$ & $G^D_{3 \times 3}$ & $\mathcal{L}_\text{stb}$ & PSNR(dB) \\
        \midrule
        \checkmark & \checkmark & \checkmark & \checkmark &  & 9.81 \\
        \rowcolor{Gray}
        \checkmark &  &  &  & \checkmark & 28.54 \\
         & \checkmark &  &  & \checkmark & 30.46 \\
        \rowcolor{Gray}
        \checkmark & \checkmark &  &  & \checkmark & 31.74 \\
        \checkmark &  & \checkmark &  & \checkmark & 32.19 \\
        \rowcolor{Gray}
         & \checkmark &  & \checkmark & \checkmark & 32.21 \\
        \checkmark & \checkmark & \checkmark & \checkmark & \checkmark & \textbf{34.08} \\
        \bottomrule
    \end{tabularx}
    \tabcspace
    \caption{
        \textbf{Model ablation study on SIDD validation set.}
        The notation of each modules follows the \figref{fig:architecture}.
        Note that feature extractor module exists when any signal-dependent module is on.
    }
    \label{tab:ablation-SIDD}
    \tabspace
\end{table}
\begin{figure}[t]
    \renewcommand{\wp}{0.24 \linewidth}
    \centering
    \subfloat[wo $\mathcal{L}_\text{stb}$]{\includegraphics[width=\wp]{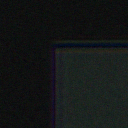}}
    \hfill
    \subfloat[wo $G^I$]{\includegraphics[width=\wp]{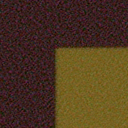}}
    \hfill
    \subfloat[all]{\includegraphics[width=\wp]{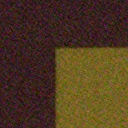}}
    \hfill
    \subfloat[GT noise]{\includegraphics[width=\wp]{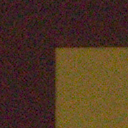}}
    \figcspace
    \caption{
        \textbf{Generated samples from model ablation study on real-world noise.}
        (a) without stabilizing loss,
        (b) without independent transforms,
        (c) with all modules and stabilizing loss,
        (d) Ground truth noisy image.
    }
    \label{fig:ablation-SIDD}
    \figspace
\end{figure}
\begin{table*}[t]
    \centering
    \renewcommand{\arraystretch}{0.95}
    \begin{tabularx}{\linewidth}{cc >{\raggedright\arraybackslash}X  cccccc }
        \toprule
        & & \multirow{2}{*}{Method} & \multicolumn{2}{c}{SIDD} && \multicolumn{2}{c}{DND} & \\
        & & & PSNR(dB) & SSIM && PSNR(dB) & SSIM  \\
        \midrule
        \multirow{4}{*}{\makecell{Non-learning based}} 
        && BM3D~\cite{BM3D} & 25.65 & 0.685 && 34.51 & 0.851 & \\
        && WNNM~\cite{WNNM} & 25.78 & 0.809 && 34.67 & 0.865 & \\
        && K-SVD~\cite{K-SVD} & 26.88 & 0.842 &&36.49 & 0.898 & \\
        && EPLL~\cite{EPLL} & 27.11 & 0.870 && 33.51 & 0.824 & \\
        \midrule
        \multirow{7}{*}{\makecell{Supervised}} 
        && TNRD~\cite{TNRD} & 24.73 & 0.643 && 33.65 & 0.831 & \\
        && DnCNN~\cite{DnCNN} & 23.66 & 0.583 && 32.43 & 0.790 & \\
        && DnCNN$^+$~\cite{DnCNN} & 35.13 & 0.896 && 37.89 & 0.932 & \\
        && CBDNet~\cite{CBDNet} & 33.28 & 0.868 && 38.05 & 0.942 & \\
        && AINDNet(R)$^*$~\cite{AINDNet} & 38.84 & 0.951 && 39.34 & 0.952 & \\
        && DIDN~\cite{DIDN} & 39.82 & 0.973 && - & - & \\
        && DANet~\cite{DANet} & 39.43 & 0.956 && 39.58 & 0.955 & \\
        \midrule
        \multirow{1}{*}{\makecell{Self-supervised}} 
        && N2V~\cite{N2V} & 27.68 & 0.668 && - & - & \\
        \midrule
        \multirow{8}{*}{\makecell{Generation-based}} 
        && GCBD~\cite{GCBD} & - & - && 35.58 & 0.922 & \\
        && UIDNet-NS~\cite{UIDNet} & 31.34 & 0.856 && - & - & \\
        && UIDNet~\cite{UIDNet} & 32.48 & 0.897 && - & - & \\
        && \textbf{C2N + DnCNN (Ours)} & \textbf{33.76} & \textbf{0.901} && \textbf{36.08} & \textbf{0.903} & \\
        && \textbf{C2N + DnCNN$^*$ (Ours)} & \textbf{34.00} & \textbf{0.907} && \textbf{36.32} & \textbf{0.908} & \\
        && \textbf{C2N + DIDN (Ours)} & \textbf{35.02} & \textbf{0.932} && \textbf{36.12} & \textbf{0.882} & \\
        && \textbf{C2N + DIDN$^*$ (Ours)} & \textbf{35.35} & \textbf{0.937} && \textbf{36.38} & \textbf{0.887} & \\
        && \textbf{C2N(SD) + DIDN$^*$ (Ours)} & - & - && \textbf{37.28} & \textbf{0.924} & \\
        \bottomrule
    \end{tabularx}
    \tabcspace
    \caption{
        \textbf{Quantitative evaluation on the SIDD and DND benchmark.}
        We adopt the two-stage pipeline which is denoted by `C2N + Denoiser'.
        $^*$ denotes the method with self-ensemble~\protect\cite{self-ens, EDSR} strategy.
        \textbf{(SD)} denotes that the C2N generator is first trained on the SIDD and then fine-tuned on the DND.
        Results of DnCNN and DnCNN$^+$ are from the model trained on synthetic noise and the model trained by ourselves on SIDD images, respectively.
    }
    \label{tab:results-real}
    \tabspace
\end{table*}
To demonstrate the validity and effectiveness of our C2N framework, we first analyze how each of the design component supports the C2N architecture in handling the synthetic noise of various properties.
\figref{fig:ablation-synth} illustrates how each modules in our C2N model can imitate two synthetic noise models, $\mathcal{P}$ and $\mathcal{S}$.
$\mathcal{P}$ stands for signal-dependent Poisson noise, and $\mathcal{S}$ stands for spatially correlated Gaussian noise of $\sigma = 50$.
To implement spatial correlation between local neighbors, we apply a $9 \times 9$ Gaussian filter to the noise map, similar to \cite{noisier2noise}.
Then, we train C2N variants with different existence of each module, e.g., $G^I_{1 \times 1}$, $G^D_{1 \times 1}$, $G^I $, and $G^D$, on BSD images corrupted by $\mathcal{P}$ and $\mathcal{S}$.
We note that the notation $G^I $ and $G^D$ refers to the combination of ($G^I_{1 \times 1}$, $G^I_{3 \times 3}$) and ($G^D_{1 \times 1}$, $G^D_{3 \times 3}$), respectively.
More details are described in our supplementary material.

The model only with signal-independent module, i.e., $G^I_{1 \times 1}$ that equally generates noise values at all locations, is not proper to synthesize signal-dependent noise.
Therefore, the output noise in \figref{g_i_1x1} does not differ in dark and bright areas compared to the ground-truth distribution in \figref{gt_p}.
By explicitly considering the signal-dependent component with the $G_{1 \times 1}^D$ module, our C2N can express such behavior of the Poisson noise as shown in \figref{g_i_1x1_g_d_1x1_on_p}.
\figref{fig:plot-ablation-Poi} further validates that our generator can synthesize signal-dependent noise, where standard deviations of the noise intensities are correlated with pixel values.
\figref{g_i_1x1_g_d_1x1_on_s} illustrates that a model with only $1 \times 1$ convolutions, i.e., $G^I_{1 \times 1} + G^D_{1 \times 1}$, cannot imitate spatially correlated noise and generates undesirable structures.
By including modules with $3 \times 3$ convolutions, i.e., $G^I_{3 \times 3}$ and $G^D_{3 \times 3}$, the proposed C2N can synthesize the desired noise distribution without unpleasing artifacts, as shown in \figref{g_i_g_d}.

We also evaluate our method on synthetic AWGN and compare the performance with supervised denoising models in \tabref{tab:result-Gau}.
The supervised model requires pairs of clean and noisy examples that correspond to the target noise level.
Otherwise, the denoising performances are degraded due to a mismatch in training and test distribution.
However, such an assumption is less practical in the real-world scenario as we may not know the exact noise levels of our target images.
The primary advantage of the proposed C2N framework is that we leverage unpaired noisy and clean images, and the exact noise level is not required to develop the following denoising method.
Our unsupervised generation-based method achieves comparable performance to the supervised denoising model on a synthetic Gaussian dataset.

\subsection{Model Analysis of C2N on Real-World Noise} \label{model_analysis-real}

\begin{figure*}[t]
    \renewcommand{\wp}{0.195 \linewidth}
    \captionsetup[subfloat]{font=small}
    \centering
    \subfloat[Noisy input]{\includegraphics[width=\wp]{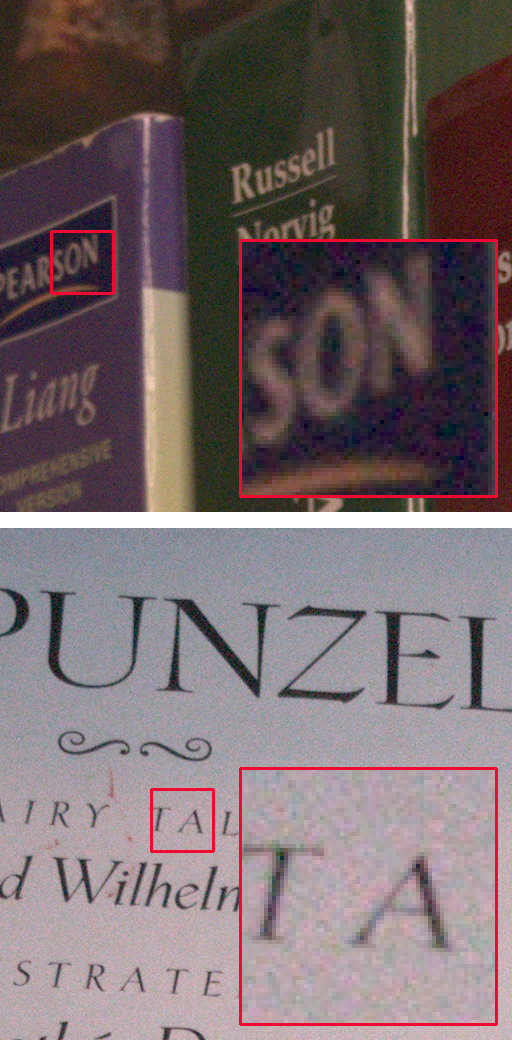}}
    \hfill
    \subfloat[BM3D \cite{BM3D}]{\includegraphics[width=\wp]{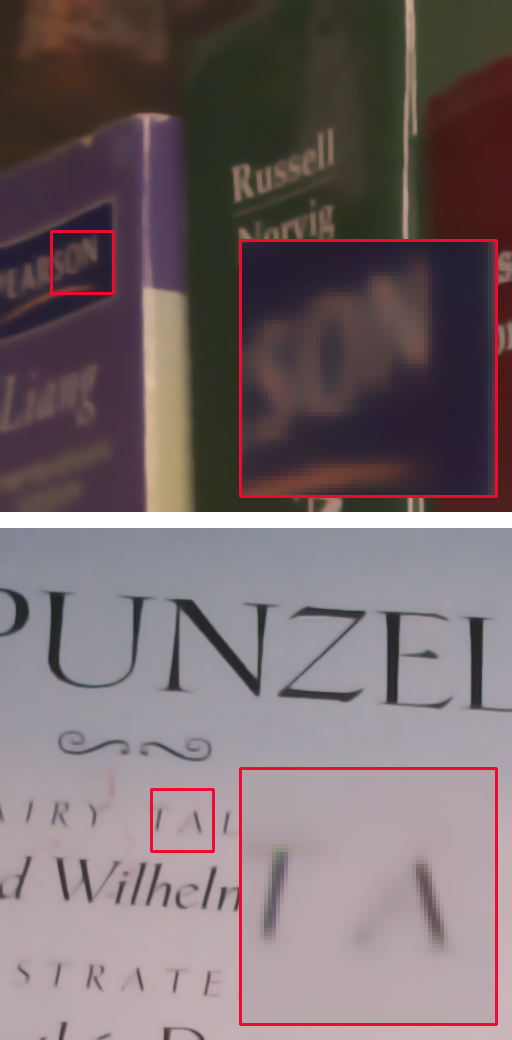}}
    \hfill
    \subfloat[UIDNet \cite{UIDNet}]{\includegraphics[width=\wp]{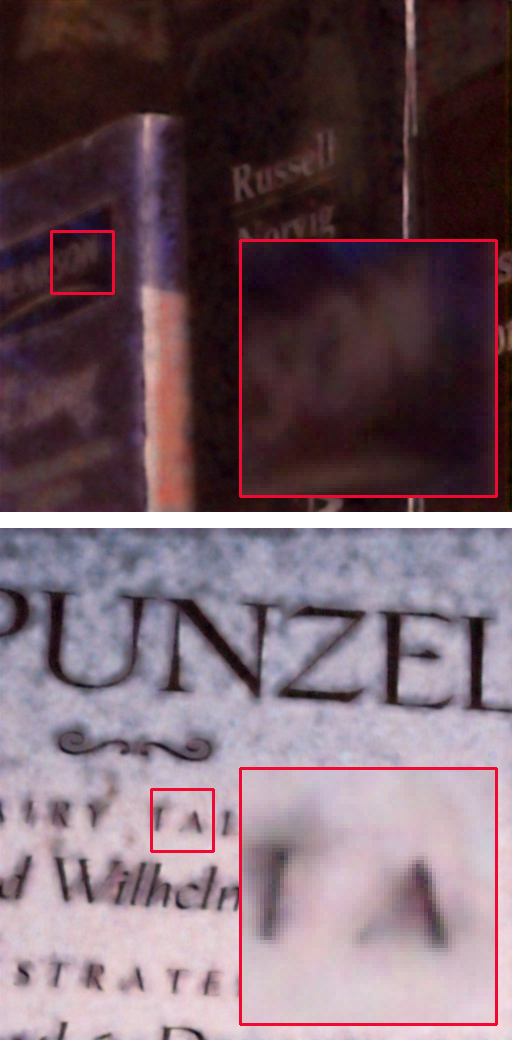}}
    \hfill
    \subfloat[\textbf{C2N + DnCNN (Ours)}]{\includegraphics[width=\wp]{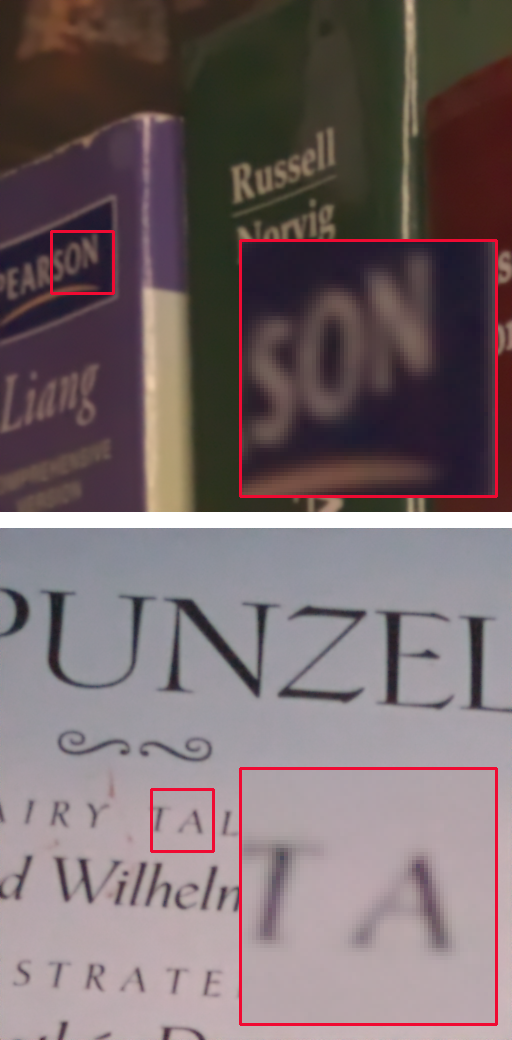}}
    \hfill
    \subfloat[DnCNN$^+$ (Supervised)]{\includegraphics[width=\wp]{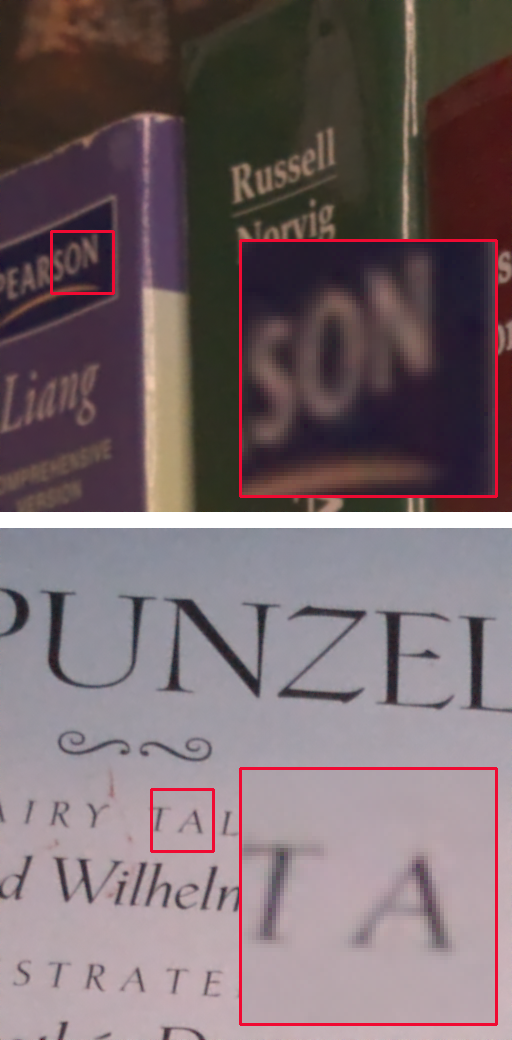}}
    \figcspace
    \caption{
    \textbf{Comparison between denoising result examples of the SIDD dataset.
    }
    (c) We use publicly released UIDNet to get the result images.
    (e) The DnCNN model is trained in a supervised manner, using clean-noisy pairs in the SIDD dataset.
    }
    \label{fig:comparison-SIDD}
    \figspace
\end{figure*}

We validate the effectiveness of each modules in our C2N framework under the real-world degradations.
\figref{fig:comparison-gen} shows that the generator in our C2N can generate realistic noise maps, whereas the existing noise generator tends to produce a particular texture.
Also, our C2N achieves the lowest KL-divergence between the generated and ground truth noise maps, among the compared methods, as show in \tabref{tab:results-KL}.
The stochastic behavior and explicit transforms of the C2N model help its output to maintain the characteristics of noise, solving the difficulties of a regular CNN functioning as a noise generator.
Note that for UIDNet experiment in \figref{fig:comparison-gen} and \tabref{tab:results-KL}, we train model with officially released code.

Also, \tabref{tab:ablation-SIDD} and \figref{fig:ablation-SIDD} show that each component of our C2N model is essential for real-world noise modeling.
Since our method learns the noise distribution without any accurate noise maps or a heuristic technique to stabilize the learning, the stabilizing loss $\mathcal{L}_\text{stb}$ plays an important role.
By comparing rows with only $1 \times 1$ module to rows combined with $3 \times 3$ module of \tabref{tab:ablation-SIDD}, we can again verify the necessity of $G^I$, $G^D$ and $3 \times 3$ convolutional transforms.
In addition, using signal-independent and dependent module together gain higher PNSR result than others.
Result shows that each single module is not enough for generating complex real-world noise.
For example, the model without $G^I$ at the 6th row of the \tabref{tab:ablation-SIDD} produces samples that differ in noise distribution, as shown in \figref{fig:ablation-SIDD}.
From the ablation study, we confirm that the entire elements of the C2N together can well simulate challenging real-world noise.

\subsection{Results of Real-World Denoising} \label{results-real}
\tabref{tab:results-real} shows the performance of our method evaluated on the DND and the SIDD benchmarks of real-world denoising in sRGB space.
We note that our method outperforms existing unsupervised methods by a large margin.
Especially, our method with C2N and DnCNN denoiser shows better results on the SIDD benchmark than the UIDNet which use similar denoising network with DnCNN model.
We also use much fewer patch data than the 520,965 noisy and clean image patches of $64 \times 64$ size used by UIDNet.
We further improve the performance of our method by using the DIDN~\cite{DIDN} as its denoising model, which has larger capacity of $\sim$217M parameters compared to that of $\sim$0.67M parameters of the DnCNN.
It shows that we can train arbitrary denoising models with C2N, unlike UIDNet with a fixed denoising backbone.

In the DND benchmark, our method also outperforms GCBD, another previous generation-based method.
Several existing methods~\cite{CBDNet, AINDNet} use rich training images of the SIDD to get high performance on the DND benchmark, even though the DND and SIDD datasets are not matching datasets.
Similarly, we also train our C2N generator on the SIDD dataset and fine-tune trained model on the DND dataset to obtain more improved performance than directly training on the DND, as shown in the last row of \tabref{tab:results-real}.

\figref{fig:comparison-SIDD} shows denoised result examples on the SIDD dataset.
Along with the quantitative results, our method shows better noise removal and detail-preserving results than the compared unsupervised methods, with quality comparable to that of the supervised model.

\section{Conclusion}
In this paper, we propose a C2N framework for practical real-world denoising which includes our novel noise generator.
By explicitly designing components of the generator considering signal-dependency and spatial correlation of real-world noise property, it successfully learn to simulate the noise distribution of the noisy images in situation of unpaired setting.
We show that each module of the model can generate noise with corresponding characteristics through the experiments on several synthetic noise and real-world noise.
Using the generated noisy and clean pairs from our generator, we train denoisers to outperform the existing methods without use of actual data pairs.
We believe our method can be a key to solve the challenging points of practical real-world denoising.

\section*{Acknowledgement}
This work was supported in part by an IITP grant funded by the Korean government [No. 2021-0-01343, Artificial Intelligence Graduate School Program (Seoul National University)], and in part by AIRS Company in Hyundai Motor Company and Kia Corporation through HMC/KIA-SNU AI Consortium Fund.

{\small
\bibliographystyle{ieee_fullname}
\bibliography{egbib}
}

\end{document}


\title{Supplementary Material \textit{for} \\ \vspace{2mm}
C2N: Practical Generative Noise Modeling for Real-World Denoising}

\author{
Geonwoon Jang\thanks{\textit{Authors contributed equally.}} ~~~~~~~~
Wooseok Lee\footnote[1] ~~~~~~~~
Sanghyun Son ~~~~~~~~
Kyoungmu Lee\\
ASRI, Department of ECE, Seoul National University\\
{\tt\small \{onwoono, adntjr4\}@gmail.com, \{thstkdgus35, kyoungmu\}@snu.ac.kr}
}

\maketitle
\renewcommand{\thetable}{S\arabic{table}}
\renewcommand{\thefigure}{S\arabic{figure}}
\renewcommand{\theequation}{S\arabic{equation}}
\renewcommand{\thesection}{S\arabic{section}}

\section{Measurement of KL-Divergence}
In Section \textcolor{blue}{4.3} of our main manuscript, the KL-divergence is evaluated with the following definition:
\begin{equation} \label{eq:kl-div}
    D_\text{KL}(P_{n} \Vert P_{\hat{n}}) = \sum_{i=0}^{255} {p_{n}(i) \log \frac{p_{n}(i)}{p_{\hat{n}}(i)}},
\end{equation}
where $P_{n}$ and $P_{\hat{n}}$ denote the distribution of the ground truth and generated noise maps, respectively.
Also, $p_{n}$ and $p_{\hat{n}}$ are the probability histogram of all noise maps from $P_{n}$ and $P_{\hat{n}}$.
The histogram $p_{n}$ is calculated as follows:
\begin{equation} \label{eq:histogram}
    p_{n}(i) = \frac{1}{mCHW} \sum_{k,c,h,w}^{m,C,H,W} \mathbf{1}_{\{n_{k,c,h,w} = i\}},
\end{equation}
where $k$, $c$, $h$, $w$ denote indices of noise maps and color channel, height, width of the corresponding image of size $H$, $W$, and $C$ and $m$ are the number of color channels and the number of the noise maps we want to evaluate, respectively.
$\mathbf{1}_{\left\{ n_{k, c, h, w} = i \right\}}$ is used as an indicator function which refers the number of pixels with noise intensity of $i$.
We note that the histogram is calculated together between different color channels.
The remaining histogram $p_{\hat{n}}(i)$ for generated noise map is also calculated in the same manner as \eqref{eq:histogram}.

\section{Implementation Details}
\begin{figure}[t]
    \centering
    \includegraphics[width=0.7\columnwidth]{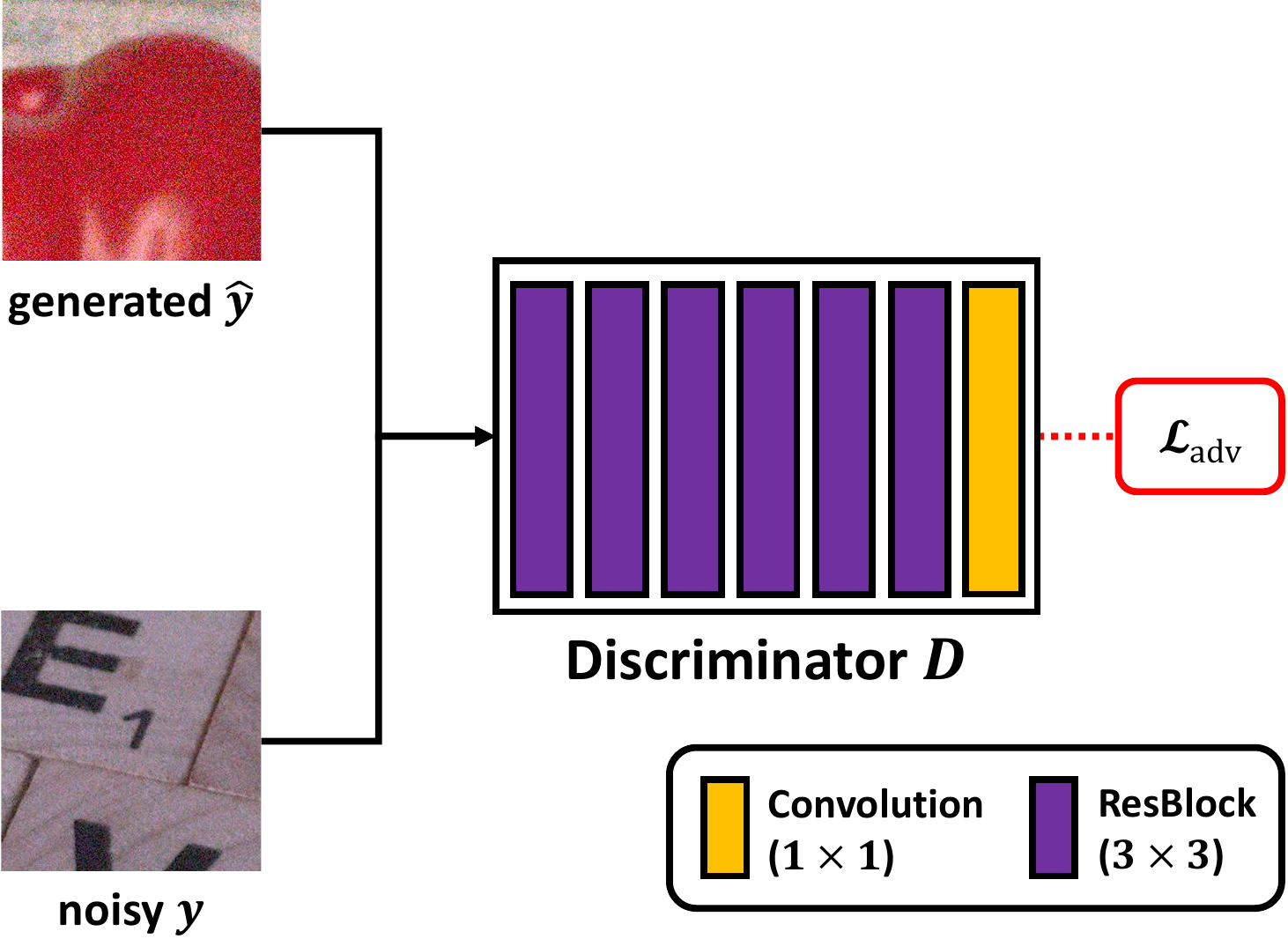}
    \figcspace
    \caption{
    \textbf{Discriminator architecture.}
    The ResBlocks, like those of the generator, are modified for low-level vision problems~\cite{EDSR}.
    }
    \label{fig:discriminator}
    \figspace
\end{figure}

\paragraph{Discriminator.}
We define the discriminator architecture as a sequence of six ResBlocks with $3 \times 3$ convolutions and the following $1 \times 1$ convolution, which reduces the number of channels to one.
\figref{fig:discriminator} shows the discriminator architecture of our C2N framework.
The output values from the discriminator is averaged across spatial dimension to indicate whether the image is real or generated one.
We set the number of channels for all $3 \times 3$ ResBlocks as 64, which is the same number as the C2N generator.

\paragraph{Self-ensemble.}
We apply the self-ensemble technique~\cite{self-ens, EDSR} to acquire final denoised results.
For a noisy image, we augment inputs by flipping and 90$^\circ$ rotations and evaluate the denoising model 8 times including the original.
We convert each output to the original geometry by the inverse transformations and average all to get the self-ensembled result.

\paragraph{Fine-tuning on the DND}
In Section \textcolor{blue}{4.4} of our main manuscript, we also report the performance of the model trained on the SIDD and then fine-tuned on the DND.
We fine-tune the C2N generator by training it for 16 more epochs on the DND dataset, with initial learning rate of $10^{-5}$ multiplied by 0.8 for every 3 epochs.

\section{Architectures in the Model Analysis}
\begin{figure*}[t]
    \renewcommand{\wp}{0.33\linewidth}
    \centering
    \subfloat{\includegraphics[width=0.35\linewidth]{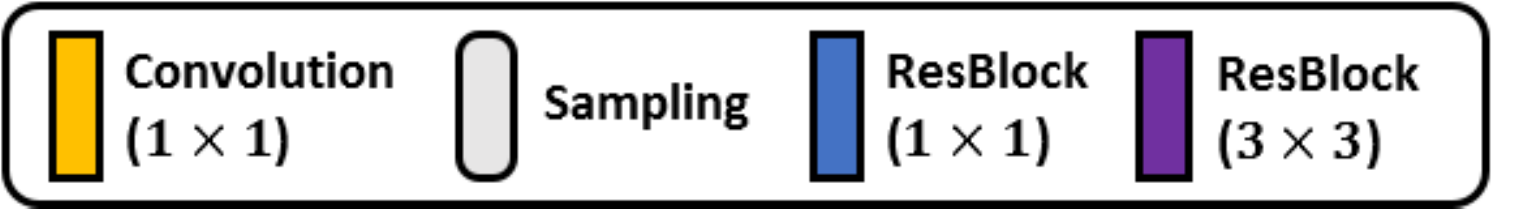}}
    \\
    \vspace{-3mm}
    \addtocounter{subfigure}{-1}
    \subfloat[$G^{I}_{1 \times 1}$ \label{synth_arch_I1}] {\includegraphics[width=\wp]{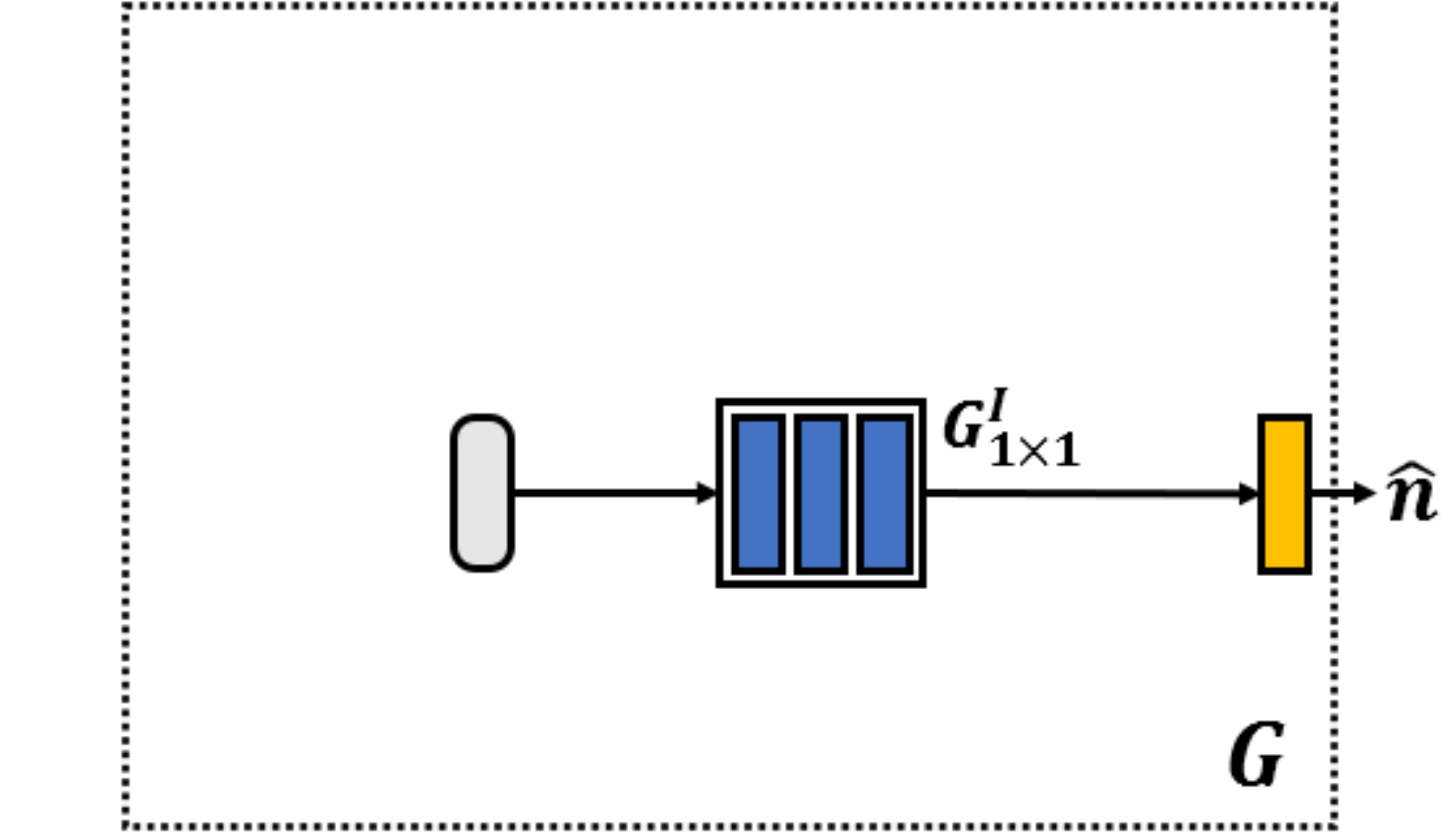}}
    \hfill
    \subfloat[$G^{I}_{1 \times 1} + G^{D}_{1 \times 1}$ \label{synth_arch_1}] {\includegraphics[width=\wp]{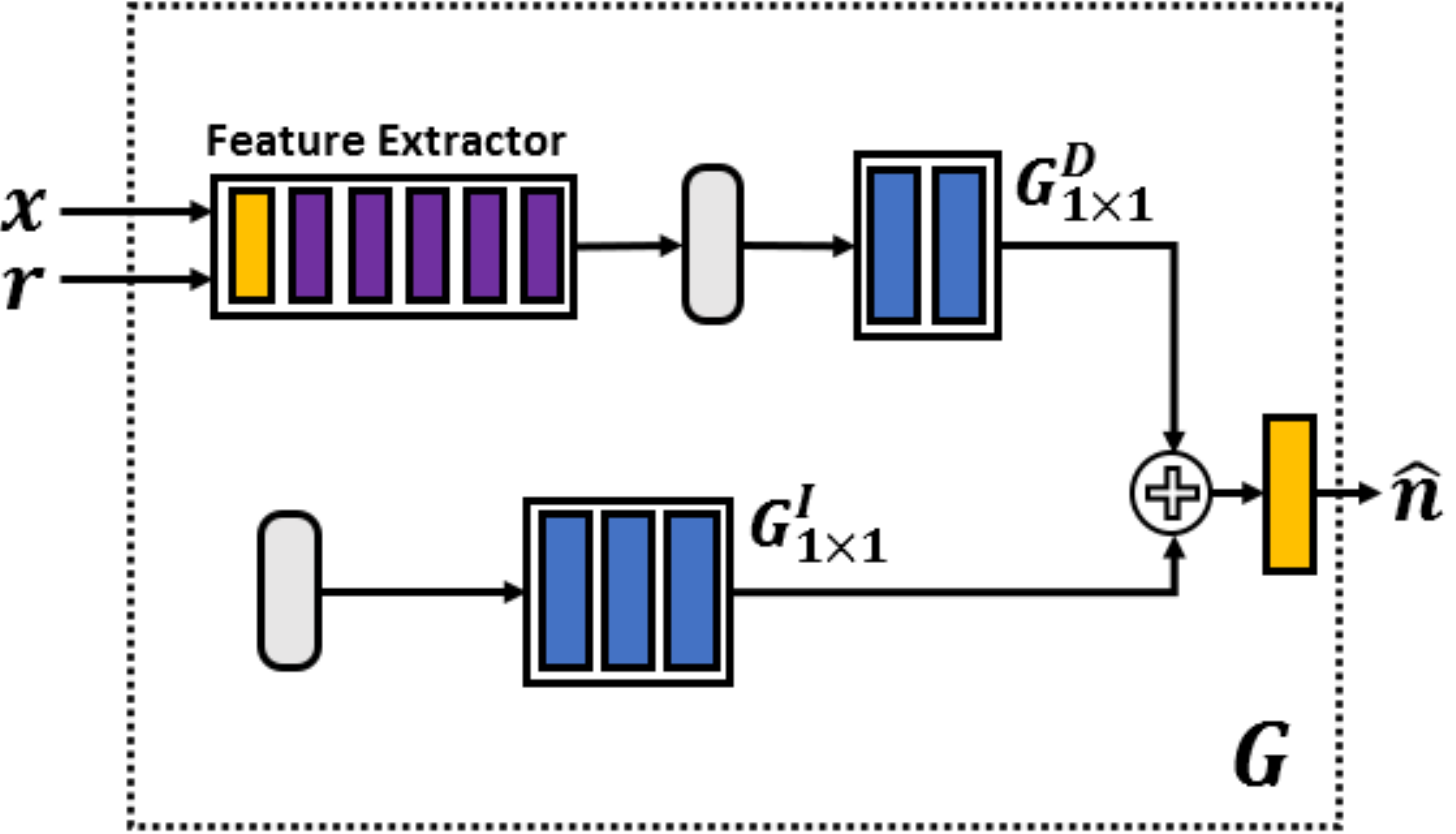}}
    \hfill
    \subfloat[$G^{I} + G^{D}$ \label{synth_arch_all}] 
    {\includegraphics[width=\wp]{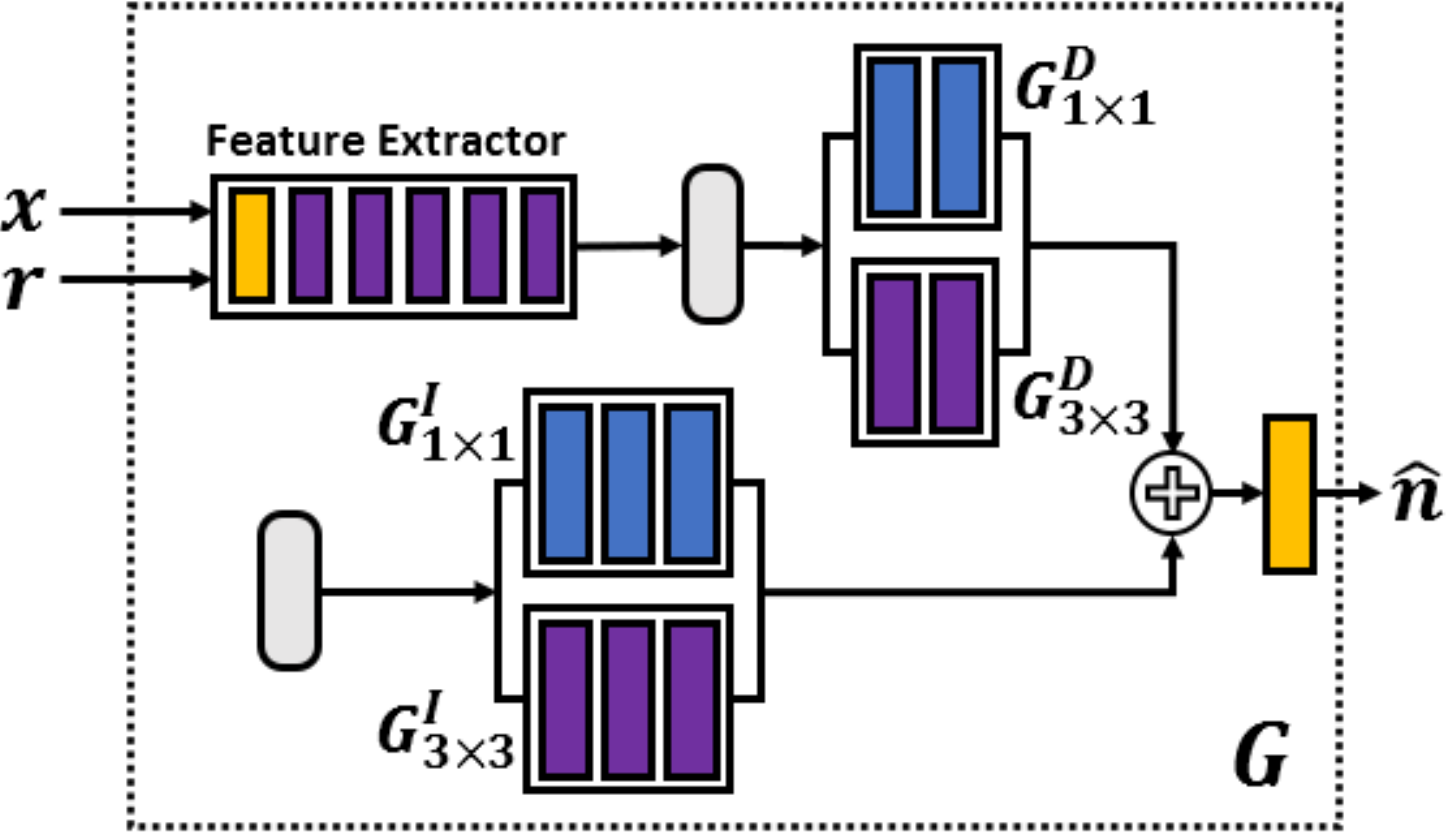}}
    \figcspace
    \caption{
    \textbf{Generator architectures in synthetic experiment.}
    (a) signal-independent and spatially uncorrelated noise generator.
    (b) signal-dependent but spatially uncorrelated noise generator.
    (c) Our whole C2N generator, which can generate signal-dependent and spatially correlated noise.
    Notations are same with Figure \textcolor{blue}{3} in our main manuscript.
    $\hat{n}$ denotes generated noise map and replication procedure of random vector $r$ is skipped for visualization.
    }
    \label{fig:synth_arch}
    \figspace
\end{figure*}

For the model ablation study on synthetic noise in Section \textcolor{blue}{4.2} of our main manuscript, we use the notations $G^{I}_{1 \times 1}$, $G^{I}_{1 \times 1} + G^{D}_{1 \times 1}$, and $G^{I} + G^{D}$ to refer the variants of our C2N.
\figref{fig:synth_arch} illustrates detailed diagrams of those variants.
The model $G^{I}_{1 \times 1}$ in \figref{synth_arch_I1} consists of a signal-independent transformation module with $1 \times 1$ convolutions that do not take clean image $x$ and random vector $r$ as input.
In contrast, the $G^{I}_{1 \times 1} + G^{D}_{1 \times 1}$ in \figref{synth_arch_1} has both modules to produce signal-independent and signal-dependent noise terms but only with $1 \times 1$ convolutions.
Lastly, the $G^{I} + G^{D}$ in \figref{synth_arch_all} consists of all modules for the proposed C2N, including $G^{I}_{3 \times 3}$ and $G^{D}_{3 \times 3}$.

\begin{figure}[t]
    \renewcommand{\wp}{0.495\linewidth}
    \renewcommand{\vs}{-3mm}
    \centering
    \subfloat[$\mathcal{P}$ (Ground Truth) \label{gt_p}]{\includegraphics[width=\wp]{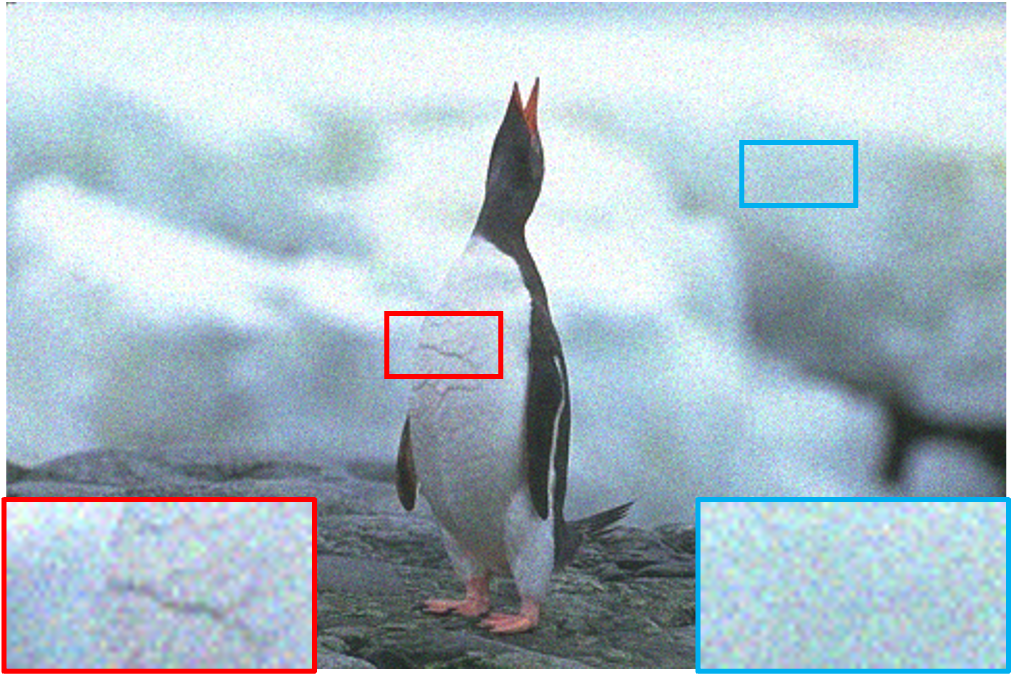}}
    \hfill
    \addtocounter{subfigure}{2}
    \subfloat[$\mathcal{S}$ (Ground Truth) \label{gt_s}]{\includegraphics[width=\wp]{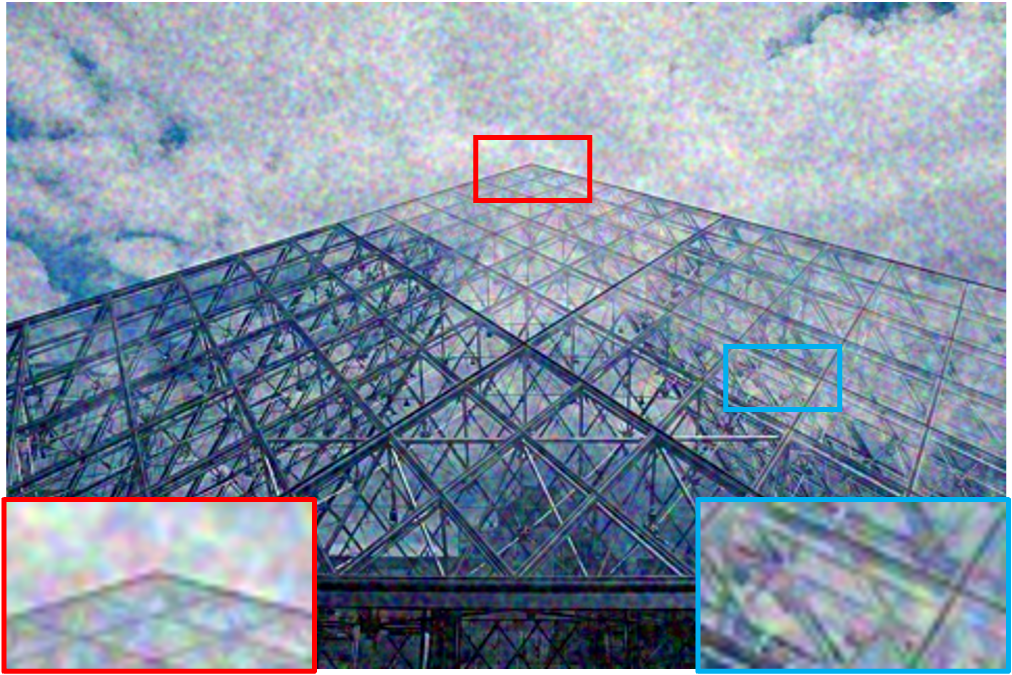}}
    \\
    \vspace{\vs}
    \addtocounter{subfigure}{-3}
    \subfloat[$G^I_{1 \times 1}$ on $\mathcal{P}$ \label{g_i_1x1}]{\includegraphics[width=\wp]{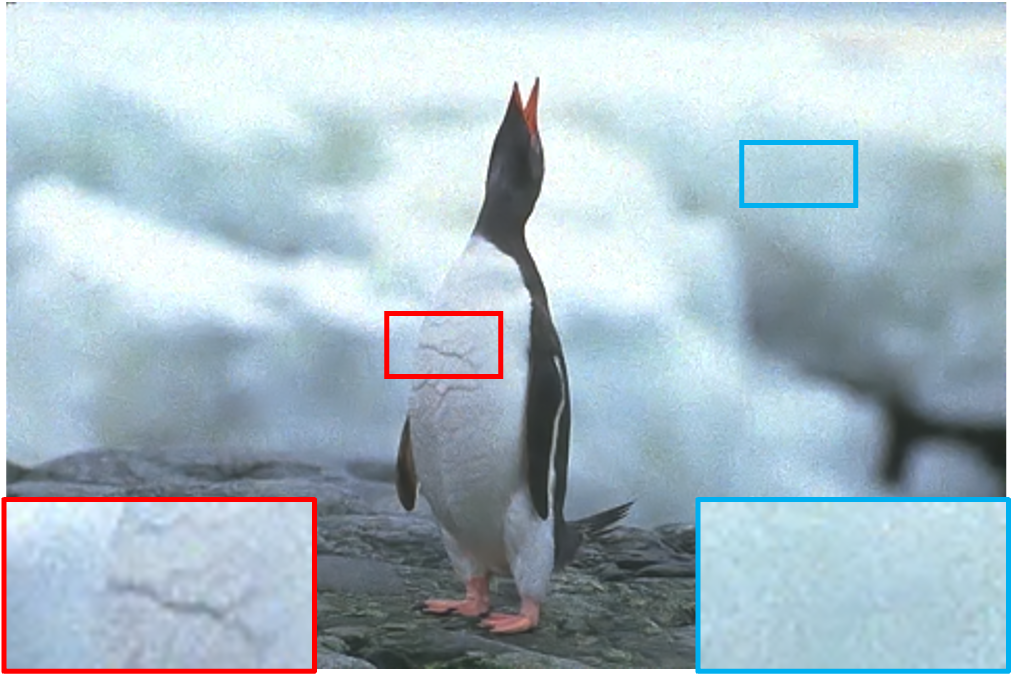}}
    \hfill
    \addtocounter{subfigure}{2}
    \subfloat[$G^I_{1 \times 1} + G^D_{1 \times 1}$ on $\mathcal{S}$ \label{g_i_1x1_g_d_1x1_on_s}]{\includegraphics[width=\wp]{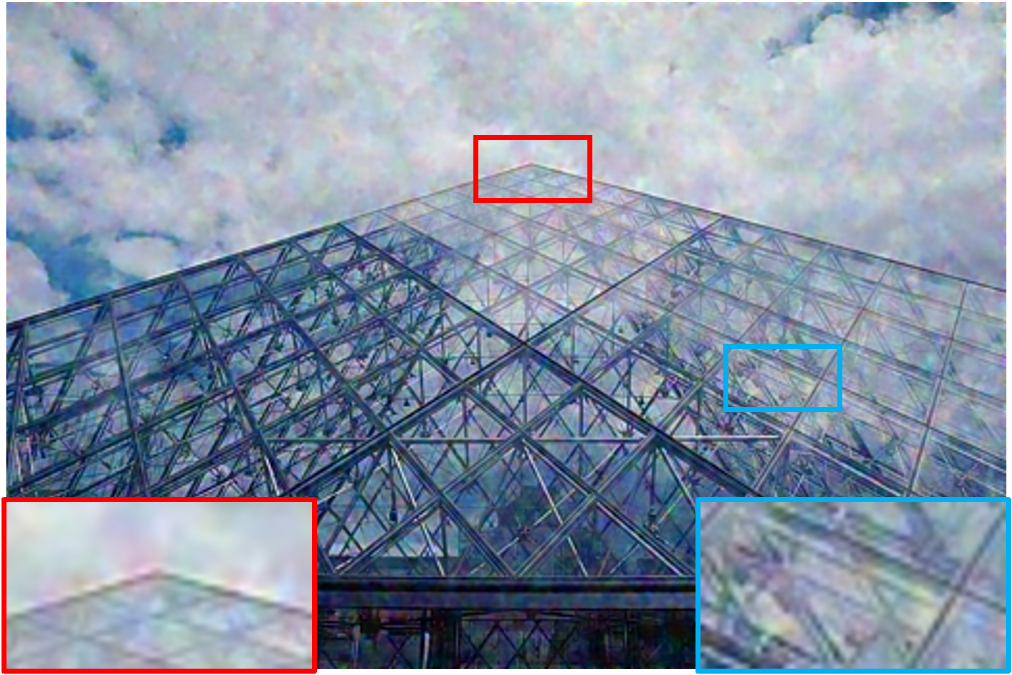}}
    \\
    \vspace{\vs}
    \addtocounter{subfigure}{-3}
    \subfloat[$G^I_{1 \times 1} + G^D_{1 \times 1}$ on $\mathcal{P}$ \label{g_i_1x1_g_d_1x1_on_p}]{\includegraphics[width=\wp]{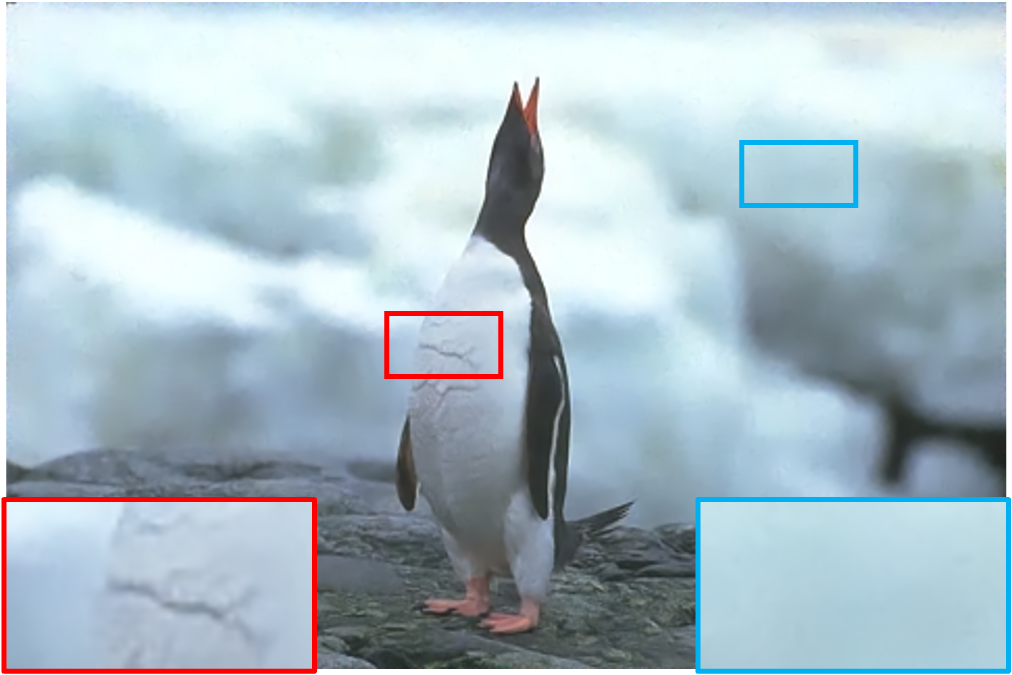}}
    \hfill
    \addtocounter{subfigure}{2}
    \subfloat[$G^I + G^D$ on $\mathcal{S}$ \label{g_i_g_d}]{\includegraphics[width=\wp]{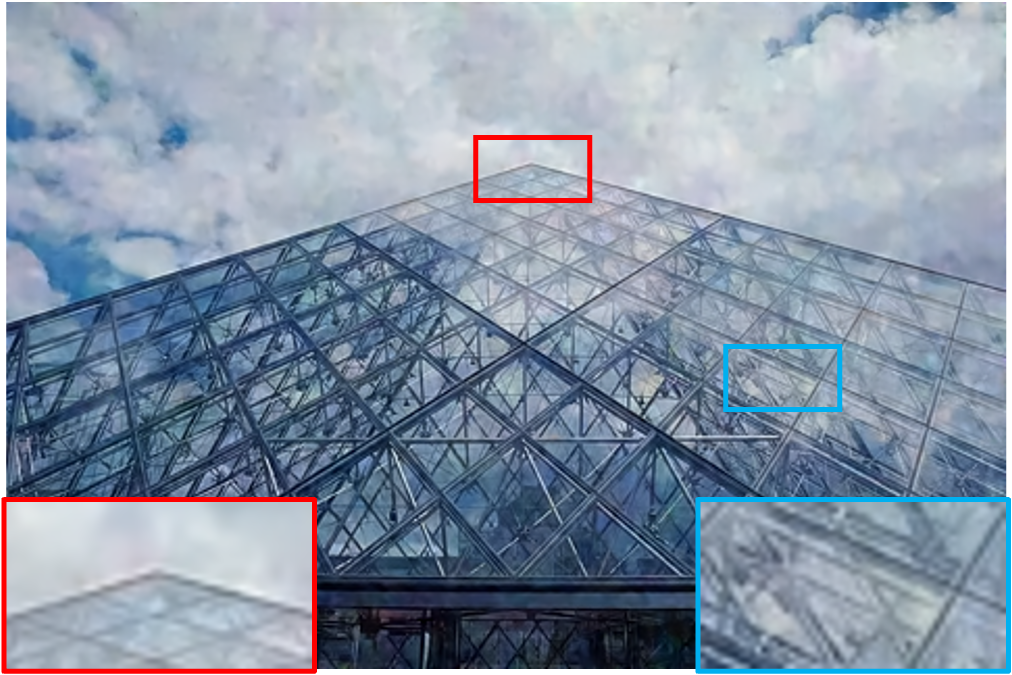}}
    \\
    \figcspace
    \caption{
    \textbf{Denoising results from model ablation study on various synthetic noise.}
    (a, d) Synthetic ground-truth noisy image of Poisson $\mathcal{P}$ and spatially correlated Gaussian noise $\mathcal{S}$.
    (b, c) Denoising results of the C2N variant trained on the Poisson noise and its following denoiser.
    (e, f) Denoising results of the C2N variant trained on the spatially correlated Gaussian noise and its following denoiser.
    }
    \label{fig:ablation-synth}
    \figspace
\end{figure}

\section{Denoising on Synthetic Noise}
The primary purpose of the model ablation study on synthetic noise in Section \textcolor{blue}{4.2} of our main manuscript is to demonstrate how our C2N can generate noise with various characteristics.
Still, we can also train the denoising networks followed by each C2N variant and evaluate their performance on various synthetic noise, as shown in \tabref{tab:ablation-synth} and \figref{fig:ablation-synth}.
$\mathcal{G}$ stands for Gaussian noise of $\sigma = 25$, $\mathcal{P}$ stands for Poisson noise $n \sim Poi(x)-x$ where $Poi(x)$ denotes the Poisson distribution similar to \cite{NoiseFlow}, and $\mathcal{S}$ stands for spatially correlated Gaussian noise.
The same notations of $\mathcal{P}$ and $\mathcal{S}$ are used in Section \textcolor{blue}{4.2} of our main manuscript.

Unlike $G^{I}_{1 \times 1} + G^{D}_{1 \times 1}$, $G^{I}_{1 \times 1}$ cannot handle signal-dependent noise level as shown in Figure \textcolor{blue}{5} of our main manuscript.
As a result, the denoiser followed by $G^{I}_{1 \times 1}$ generator in \figref{g_i_1x1} is not appropriate to remove non-uniform Poisson noise.
Meanwhile, in \figref{g_i_1x1_g_d_1x1_on_s} the denoiser followed by $G^{I}_{1 \times 1} + G^{D}_{1 \times 1}$ outputs images that still contain noise term of $\mathcal{S}$, since the $G^{I}_{1 \times 1} + G^{D}_{1 \times 1}$ tends to generate artifacts instead of spatially correlated noise, as shown in Figure \textcolor{blue}{4} of our main manuscript.

\begin{table}[t]
    \centering
    \renewcommand{\arraystretch}{0.9}
    \setlength{\belowcaptionskip}{4pt}
    \begin{tabularx}{\linewidth}{p{2.4cm} >{\centering\arraybackslash}X>{\centering\arraybackslash}X>{\centering\arraybackslash}X}
        \toprule
        & \multicolumn{3}{c}{Test Noise Type} \\
        C2N Model & $\mathcal{G}$ & $\mathcal{P}$ & $\mathcal{S}$ \\
        \midrule
        $G^{I}_{1 \times 1}$ & 30.69 & 35.21 & 26.09 \\
        $G^{I}_{1 \times 1} + G^{D}_{1 \times 1}$ & 30.66 & \textbf{35.80} & 28.91 \\
        $G^{I} + G^{D}$ & 30.40 & 35.23 & \textbf{31.03} \\
        \bottomrule
    \end{tabularx}
    \tabcspace
    \caption{
    \textbf{Denoising performance on various synthetic noise.}
    PSNR(dB) is calculated on the CBSD68 dataset.
    }
    \label{tab:ablation-synth}
    \tabspace
\end{table}

\section{Visualizing Generated Noise}
\paragraph{Comparison between generated samples and the real-world noise.}
\begin{figure*}[t]
    \renewcommand{\wp}{0.195\linewidth}
    \centering
    \subfloat[Clean]{\includegraphics[width=\wp]{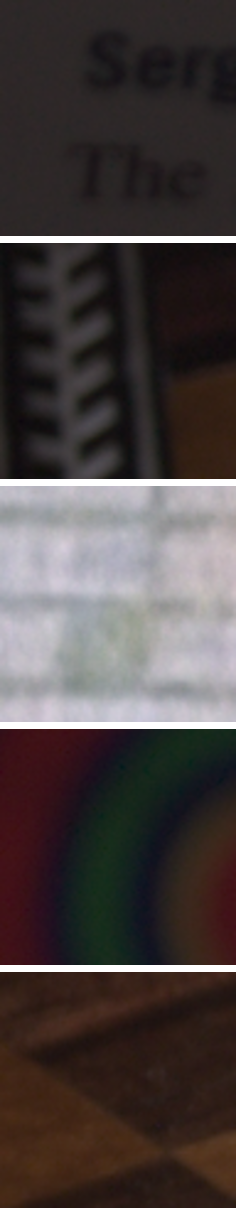}}
    \hfill
    \subfloat[GT]{\includegraphics[width=\wp]{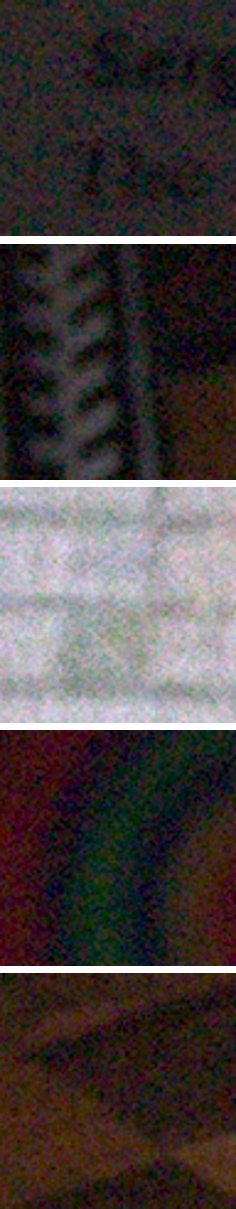}}
    \hfill
    \subfloat[GT res]{\includegraphics[width=\wp]{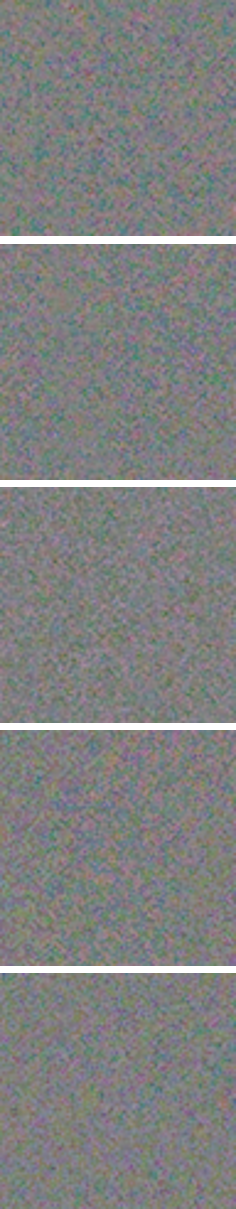}}
    \hfill
    \subfloat[C2N]{\includegraphics[width=\wp]{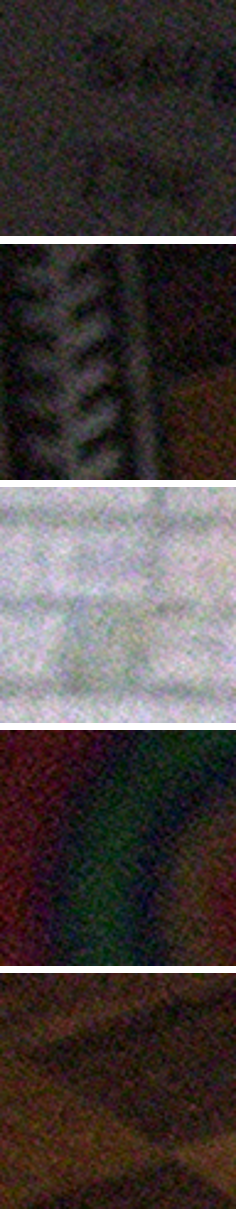}}
    \hfill
    \subfloat[C2N res]{\includegraphics[width=\wp]{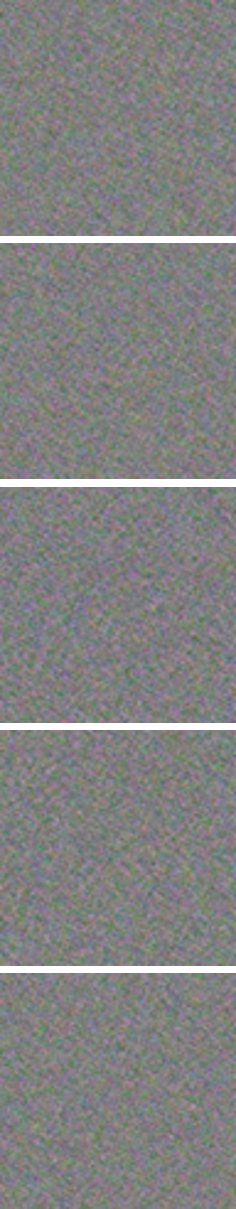}}

    \figcspace
    \caption{
    \textbf{Examples of ground truth noisy image and generated image from our C2N.}
    (a) Clean image, (b) Ground truth noisy image and (c) its residual noise map,
    (d) Generated noisy image from the proposed C2N and (e) its residual noise map.
    Best with zoomed.
    }
    \label{fig:sup-more}
    \figspace
\end{figure*}
\figref{fig:sup-more} visually compares more pseudo-noisy samples generated by our C2N and ground-truth noise maps.
The proposed C2N can synthesize samples that closely resemble ground-truth noise without significant artifacts.

\paragraph{Latent space interpolation.}
We also provide a qualitative study on the effects of $r$, the input random vector of the generator.
\begin{figure*}[t]
    \renewcommand{\wp}{0.165\linewidth}
    \captionsetup[subfloat]{labelformat=empty}
    \centering
        \subfloat[$\lambda=0.0$]{\includegraphics[width=\wp]{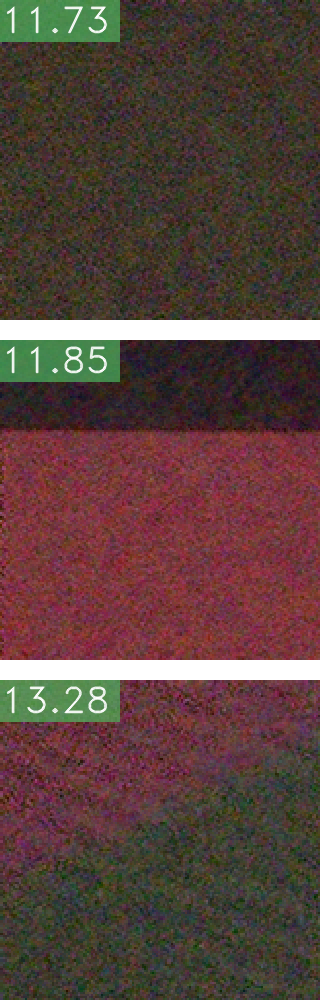}}
        \hfill
        \subfloat[$\lambda=0.2$]{\includegraphics[width=\wp]{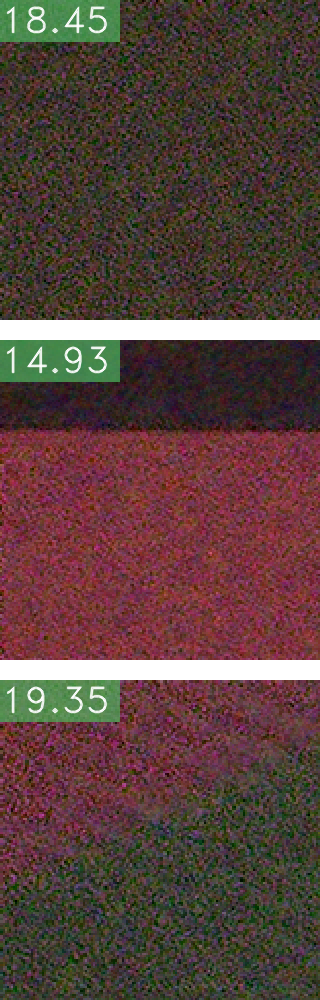}}
        \hfill
        \subfloat[$\lambda=0.4$]{\includegraphics[width=\wp]{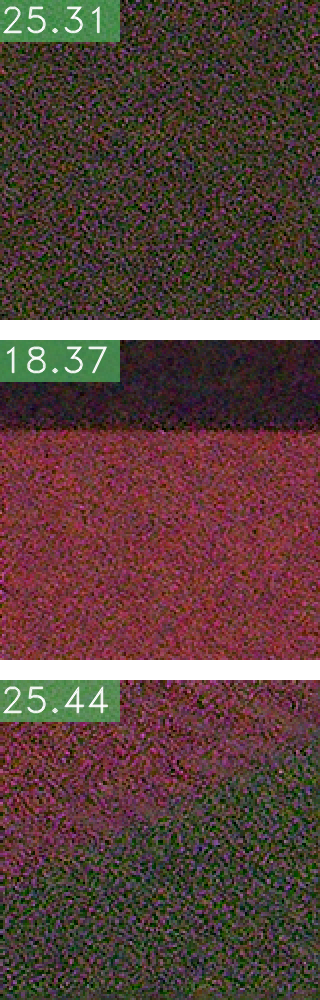}}
        \hfill
        \subfloat[$\lambda=0.6$]{\includegraphics[width=\wp]{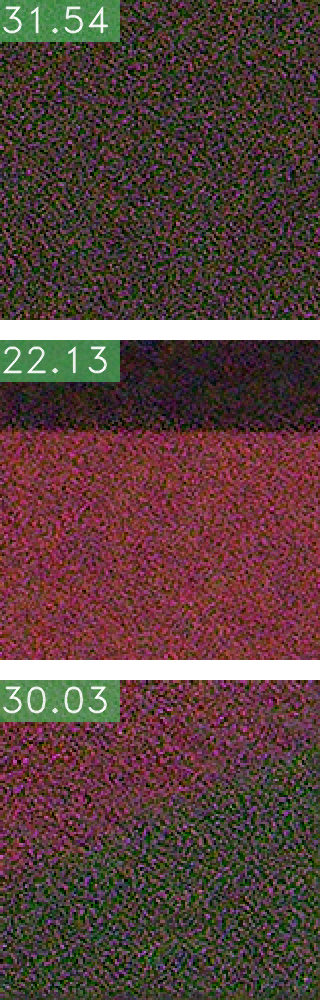}}
        \hfill
        \subfloat[$\lambda=0.8$]{\includegraphics[width=\wp]{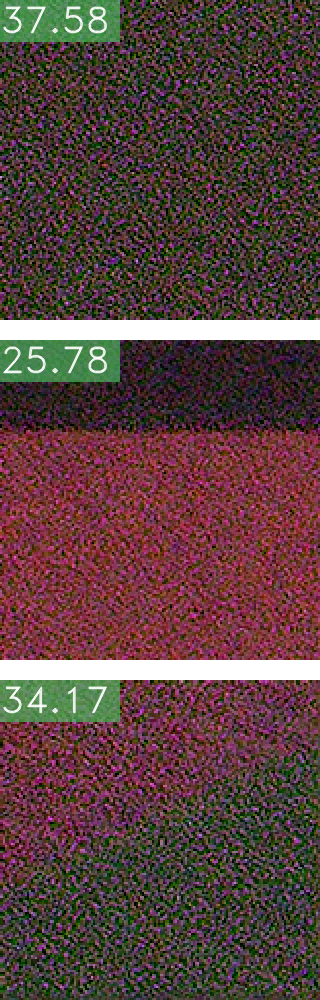}}
        \hfill
        \subfloat[$\lambda=1.0$]{\includegraphics[width=\wp]{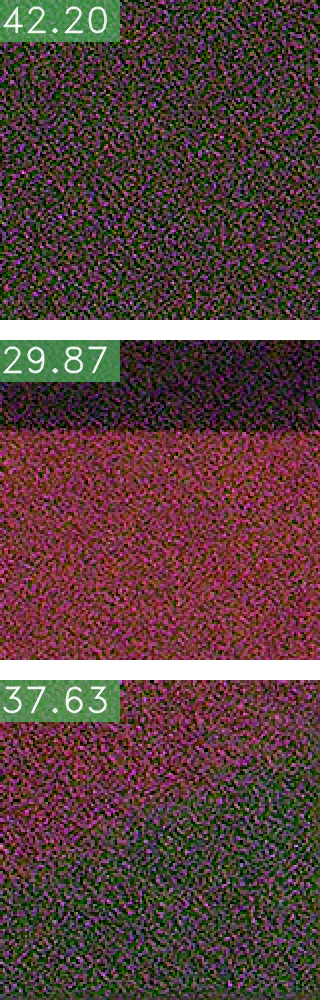}}
    \figcspace
    \caption{
    \textbf{Image generation with interpolated $r$ .}
    For the same clean image patch, we interpolate two different $r$ vectors with a factor $\lambda$ and obtain the resulting images.
    Each column is generated using same vector $r$.
    The standard deviation of each residual noise map is displayed in each image.
    Best with zoomed.
    }
    \label{fig:sup-interp}
    \figspace
\end{figure*}
We sample two $r$ vectors which generate low and high level noise and visualize the generated images with interpolated $r$ as following equation, $r=(1-\lambda)r_1+ \lambda r_2$. Here, $r_1$ and $r_2$ are the two sampled vectors and $\lambda \in \left[ 0, 1 \right]$ is the interpolation factor.
The qualitative results in \figref{fig:sup-interp} illustrates that the random vector determines the property of synthesized noise in our C2N framework.
In other words, our approach can learn to generate real-world noise that corresponds to varying conditions, such as strong or weak noise from various camera types in the SIDD~\cite{SIDD}.

\section{Practical Data Constraints in C2N} \label{sec:results-SG}

To apply the generative noise modeling methods in practical situations, two kinds of data constraints should be further considered for better usability.
First, due to several physical limitations~\cite{N2N}, it is not feasible to capture an ideal clean image from the wild.
Rather, a long sequence of aligned noisy images must be captured beforehand~\cite{SIDD} to synthesize the pseudo-clean reference.
Thus, only a few clean images are available from the same scene distribution of the noisy images in a real situation.
Secondly, once the noise generator is trained on the desired noisy image distribution $P_{N}$ and clean image distribution $P_{C}$, it should be able to produce pseudo-noisy images paired to any clean image $x$ from different clean image distribution $P'_{C}$ to train a denoising model.
Various real-world noisy image datasets have scenes that differ in many points, such as types and scales of the contents or illumination, making a model hard to learn the noise distribution distinct from the domain of scenes.

The existing generative noise modeling methods~\cite{GCBD, NoiseFlow, UIDNet} used a large number of samples in $P_{C}$, and assumed the external clean image distribution $P'_{C}$ for training denoising network to be the same as $P_{C}$.
Such a setting is possible only if a sufficiently large noisy and clean image dataset is given, which is not a practical situation.
We examine whether our method can maintain its usability under this problem that have not been explored before.

\begin{table}[t]
    \centering
    \renewcommand{\arraystretch}{1.0}
    \begin{tabularx}{\linewidth}{cccc >{\centering\arraybackslash}X >{\centering\arraybackslash}X}
        \toprule
        Number of Samples $\sim P_{C}$ & $P'_{C}$ & PSNR(dB) & SSIM \\
        \midrule
        \multirow{4}{*}{\makecell{36K\\(100\%)}} & S & 34.08 & 0.909 \\
        & D & 31.72 & 0.826 \\
        & B & 31.74 & 0.825 \\
        & U & 31.32 & 0.803 \\
        \midrule
        \multirow{4}{*}{\makecell{18K\\(50\%)}} & S & 33.53 & 0.882 \\
        & D & 30.68 & 0.760 \\
        & B & 29.96 & 0.742 \\
        & U & 29.72 & 0.741 \\
        \midrule
        \multirow{4}{*}{\makecell{720\\(2\%)}} & S & 31.98 & 0.847 \\
        & D & 29.36 & 0.745 \\
        & B & 29.27 & 0.735 \\
        & U & 29.21 & 0.738 \\
        \midrule
        \multirow{4}{*}{\makecell{360\\(1\%)}} & S & 31.84 & 0.849 \\
        & D & 29.35 & 0.740 \\
        & B & 29.08 & 0.733 \\
        & U & 29.24 & 0.739 \\
        \bottomrule
    \end{tabularx}
    \tabcspace
    \caption{
        \textbf{Denoising performance of our C2N under data constraints.}
        We use the Urban100~\protect\cite{Urban100} dataset along with the other datasets mentioned in main manuscript as the samples of $P'_{C}$.
        S, D, B, U denote the SIDD, the DIV2K high-resolution images, the BSD traning images, and the Urban100, respectively.
        $P_{C}$ is fixed to S for all experiments.
        Evaluation is done on the SIDD validation set.
    }
    \label{tab:results-data_constraint}
    \tabspace
\end{table}

\tabref{tab:results-data_constraint} shows that our method fairly preserves its performance without collapsing under two data constraints, (1) where not enough clean images in $P_{C}$ are given to train the noise generator, (2) where the clean images from different scene distribution $P'_{C}$ are used to train the following denoising model.
Our method already uses surprisingly small amount of samples for training the C2N model, compared to $\sim$500K image patches of $64 \times 64$ size used in the previous generative noise modeling methods \cite{NoiseFlow, UIDNet}.
The C2N model trained with much smaller amount of clean images in $P_{C}$ still shows performance comparable to previous unsupervised denoising methods.
Also for the case of $P'_{C}$ to be different to $P_{C}$, our C2N can still train the following denoising model with its generated pseudo-noisy data.
Although our method shows generalization ability in these situations with data constraints, we believe that further improvement to resolve such problems entirely would be an essential topic to handle in future work.

{\small
\bibliographystyle{ieee_fullname}
\bibliography{supplement}
}